\documentclass[]{spie}  

\usepackage{amsmath,amsfonts,amssymb}
\usepackage{graphicx}
\usepackage[colorlinks=true, allcolors=blue]{hyperref}

\title{MentalGame: Predicting Personality-Job Fitness for Software Developers Using Multi-Genre Games and Machine Learning Approaches}

\author[a]{Soroush Elyasi}
\author[a]{Arya VarastehNezhad}
\author[a]{Fattaneh Taghiyareh* }

\affil[a]{Department of Computer Engineering, University of Tehran}

\begin{document} 

\pagenumbering{arabic}

\setcounter{topnumber}{100}
\setcounter{bottomnumber}{100}
\setcounter{totalnumber}{100}

\renewcommand{\topfraction}{0.95}
\renewcommand{\bottomfraction}{0.95}
\renewcommand{\textfraction}{0.05}
\renewcommand{\floatpagefraction}{0.9}

\maketitle

\begin{abstract}
Personality assessment in career guidance and personnel selection traditionally relies on self-report questionnaires, which are susceptible to response bias, fatigue, and intentional distortion. Game-based assessment offers a promising alternative by capturing implicit behavioral signals during gameplay. This study proposes a multi-genre serious-game framework combined with machine-learning techniques to predict suitability for software development roles. Developer-relevant personality and behavioral traits were identified through a systematic literature review and an empirical study of professional software engineers. A custom mobile game was designed to elicit behaviors related to problem solving, planning, adaptability, persistence, time management, and information seeking. Fine-grained gameplay event data were collected and analyzed using a two-phase modeling strategy where suitability was predicted exclusively from gameplay-derived behavioral features. Results show that our model achieved up to 97\% precision and 94\% accuracy. Behavioral analysis revealed that proper candidates exhibited distinct gameplay patterns, such as more wins in puzzle-based games, more side challenges, navigating menus more frequently, and exhibiting fewer pauses, retries, and surrender actions. These findings demonstrate that implicit behavioral traces captured during gameplay is promising in predicting software-development suitability without explicit personality testing, supporting serious games as a scalable, engaging, and less biased alternative for career assessment.
\end{abstract}


\section{INTRODUCTION}
\label{sec:intro}  

Personality assessment plays a critical role in a wide range of contemporary decision-making processes. Insights derived from personality data are routinely used in credit-risk assessment, employee recruitment, prediction of job satisfaction and task performance, team composition, and educational guidance \cite{landers2022game, stuanescu2020game, amin2025comparative, acuna2009personality, varastehnezhad2025jungian, rodriguez2013using}. In many of these contexts, researchers analyse diverse forms of behavioral data, from social-media traces to digital activity logs, to discover patterns that can inform reliable predictive models \cite{vinciarelli2014survey, kedar2015automatic}. However, despite the widespread use of personality questionnaires as the dominant assessment instrument, these tools suffer from several structural limitations. Self-report bias, fatigue, misinterpretation of items, and intentional impression-management undermine the accuracy and fairness of questionnaire-based evaluations \cite{pittenger2005cautionary, Elyasi2023mbtiJ, randall2017validity, mccord2019game, schweiger1985measuring, harman2025gamified}. Consequently, identifying alternative methods capable of capturing authentic behavioral tendencies and underlying psychological paradigms has become an important research goal, motivating researchers to explore new approaches \cite{motlagh2025ai, varastehnezhad2025ai, naz2025machine}.

Computer games represent one of the most promising avenues for such alternative assessment \cite{landers2022game, mccord2019game, Elyasi2023mbtiC}. Games are enjoyable, engaging, and typically free from the psychological pressure associated with formal testing environments \cite{leutner2023game}. Players interact spontaneously, make unconstrained decisions, and display behavioral patterns that often map more directly onto their true preferences, abilities, and dispositions \cite{Elyasi2023mbtiC}. The capacity of games to record fine-grained event logs, movement patterns, decision points, response times, exploration choices, unnecessary or optional actions, and hesitation, provides an exceptionally rich behavioral dataset \cite{harman2025gamified, elyasi2025play, van2011test, zulkifly2019personality, haizel2021personality}. Because these traces are produced subconsciously, game-based assessment reduces many of the problems inherent in traditional self-report formats. This motivates a deeper investigation into the potential of games as a reliable medium for personality and behavioral analysis.

Much of this prior work has examined the connections between player decisions and personality traits in widely known commercial games such as Skyrim, Battlefield III, League of Legends, World of Warcraft, Minecraft, and Dota II \cite{Elyasi2023mbtiC, graham2013personality, wohn2013virtual, feldman2017exploring}. Other studies have designed simple, often text-based or puzzle-based games that embed personality-driven mechanics directly into gameplay. Collectively, these investigations show strong evidence that game behaviour correlates with various psychological constructs, including tendencies toward cooperation, impulsivity, problem-solving strategies, emotional stability, and even vulnerability to gaming addiction \cite{wohn2013virtual, muller2020big, vollmer2014computer, bean2016video, seok2015predicting, worth2015dimensions}.

Despite this progress, significant gaps remain. First, most existing studies rely on the Five-Factor Model (Big Five) as the underlying psychological framework, while the Myers-Briggs Type Indicator (MBTI), although widely used in organisational and career-placement contexts, has received considerably less attention \cite{pittenger2005cautionary, cruz2015forty, furnham1996big}. The dominance of Big Five-based research and the relative scarcity of MBTI-based studies after 2018-2023 (Table 1) indicate a clear imbalance in the literature. Second, although games have been studied as tools for personality inference, very few works use game data for practical job-recommendation or personnel-selection tasks, and even fewer address specific professions. Research that targets real-world applications, such as identifying candidates suited for particular roles, remains limited in both scope and methodological diversity. Third, existing studies rarely combine personality questionnaires with additional behavioral traits that may be specific to a given profession. This creates an important opportunity to design more comprehensive and domain-aware assessment frameworks. Overall, the contributions of this research are threefold.

\begin{enumerate}
  \item A comprehensive review of game-based personality assessment literature is provided, highlighting methodological patterns, questionnaire selection, and the evolution of research trends since 2010.
  \item A novel game-based assessment framework is proposed that combines MBTI with complementary behavioral traits specifically associated with software developers, bridging an identified gap in existing research.

  \item An operational predictive model is developed using gameplay event data, demonstrating that behavioral traces captured during gameplay can be used to identify individuals with potential aptitude for software-development roles. 
\end{enumerate}

Figure 1 represents the graphical abstract of this research.

\begin{figure}[ht]
\centering
\includegraphics[width=\linewidth]{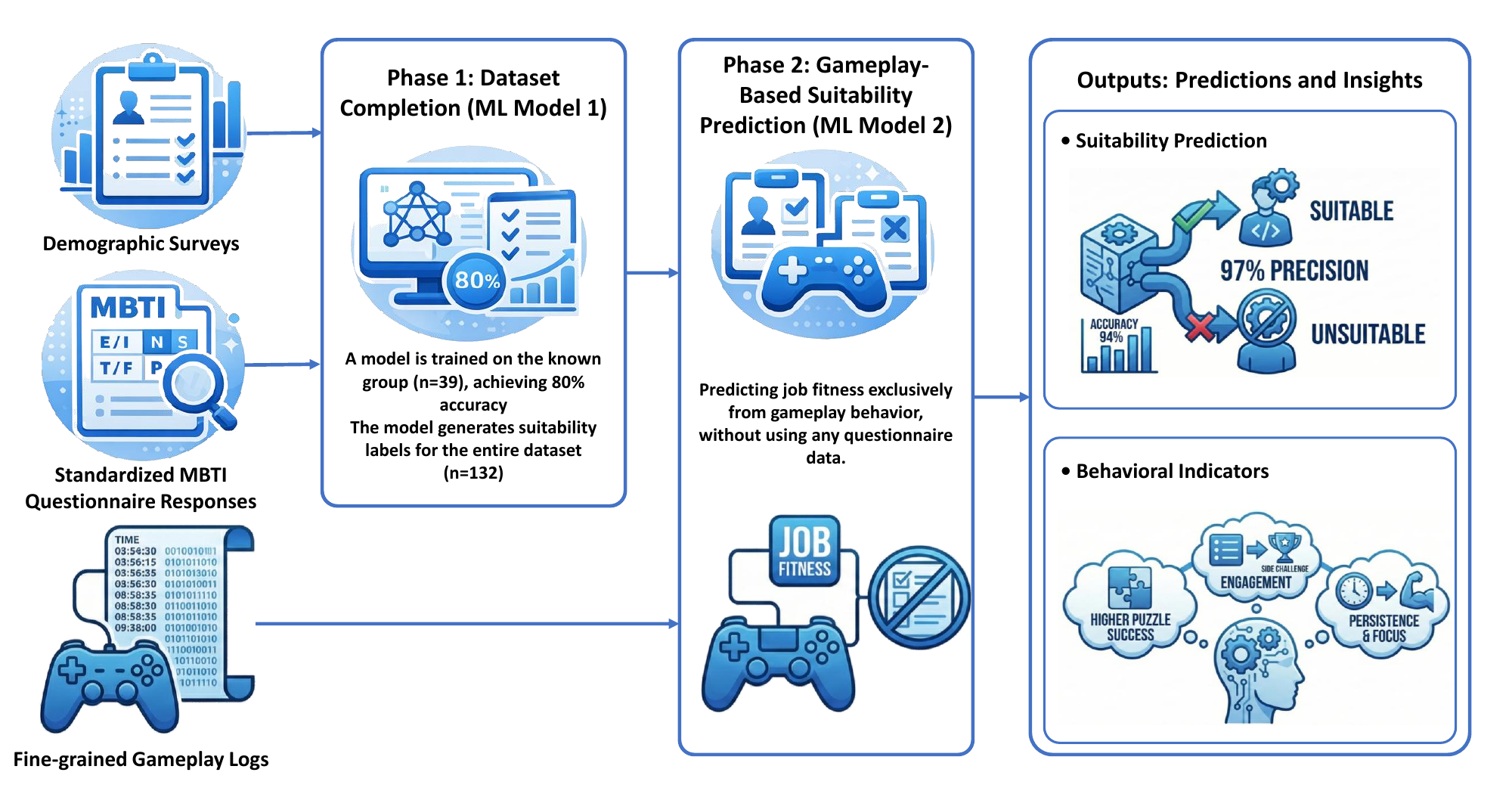}
\caption{Graphical abstract of the research}
\label{fig:pdf_figure}
\end{figure}

\section{Related Work}
Psychological tests have long been used as scientific tools for assessing individual behavior types and cognitive abilities. Over time, these tests have been standardized and are commonly administered through questionnaires. However, questionnaire-based assessments suffer from several well-known limitations, including respondent fatigue, response bias, anxiety, intentional or unintentional incorrect answers, and socially desirable responding \cite{capretz2002software}. Despite these drawbacks, such assessments remain popular among employers for personnel selection and team formation \cite{barends2023construct}. They are also widely applied in diagnosing psychological disorders and in designing educational and psychological improvement programs \cite{mccord2019game, weargeneralizing}.

A fundamental challenge of questionnaire-based assessments is their reliance on self-reporting, which inherently introduces bias. Questions may be misunderstood or misread, and questionnaires are often perceived as boring. Empirical studies show that when individuals engage in tasks they find tedious, their concentration decreases, leading to random or unrealistic responses \cite{mccord2019game, capretz2002software}. These shortcomings motivate the exploration of alternative assessment methods.

Gamification and game-based assessment approaches aim to mitigate the limitations of traditional questionnaires by embedding assessment mechanisms within gameplay \cite{mccord2019game}. By collecting behavioral data implicitly during gameplay, these methods can reduce bias, increase engagement, and yield more reliable results \cite{mccord2019game, capretz2002software}.

Numerous studies have demonstrated meaningful relationships between in-game behaviors, player choices, performance metrics, and personality traits \cite{gao2023research, quwaider2023shooter}. Research on popular games such as Skyrim, Battlefield III, League of Legends, World of Warcraft, Minecraft, and Dota II has confirmed strong associations between personality traits and in-game behavior \cite{gao2023research}. Other studies have focused on designing custom games aligned with psychological frameworks, particularly the Five-Factor Model (Big Five), and have successfully identified significant correlations between behavioral traits, problem-solving strategies, and gameplay patterns \cite{zulkifly2019personality, haizel2021personality, van2011games}.

Zulkifly developed a game designed to analyze players’ problem-solving approaches, map exploration behavior, pause duration, movement patterns (essential and non-essential), and infer personality traits from these behaviors \cite{zulkifly2019personality}. Another study employed a text-based game where each question offered three response levels (low, medium, high). This approach aimed to model the Big Five personality traits using textual gameplay and achieved higher correspondence with questionnaire results compared to traditional methods \cite{mccord2019game}. Haizel et al. introduced a role-playing game featuring a short fantasy-themed narrative with 12 embedded decision points. Although the number of choices was relatively small, the game, on average, yielded more accurate personality estimates than traditional post-game questionnaires \cite{haizel2021personality}.

Another study combined the NEO-PI questionnaire with a custom-designed game involving 44 players and 275 variables to identify five personality traits. Results demonstrated a strong correlation between game-based outputs and questionnaire-based assessments \cite{van2011games}. Another Unity-based game was tested on 30 players without revealing the assessment purpose to participants. The game lasted longer than a traditional questionnaire and used the NEO-PI for validation. Results indicated a direct relationship between player behavior and questionnaire outcomes, with higher participant satisfaction compared to traditional assessments \cite{afroza2021development}.

Gao et al. investigated personality recognition in pre-existing online games. In their first study, Clash of Kings was analyzed using K-means clustering followed by statistical correlation analysis. The HEXACO personality framework was employed, yielding the following findings \cite{gao2023research}:

\begin{itemize}
  \item Players with high extraversion exhibited more aggressive behaviors.
  \item Conscientiousness showed a negative correlation with in-game deaths.
  \item Emotionality demonstrated extreme behavioral effects, with high or low scores correlating with killing and attacking behaviors.
  \item Honesty-Humility, Agreeableness, and Openness did not significantly predict in-game behavior.
\end{itemize}

Another study focused on shooter games and used the Five-Factor Model with machine-learning techniques to predict personality traits. A custom game named Protector was developed, and Artificial Neural Networks achieved the highest accuracy, while SVR and KNNR yielded the lowest prediction errors \cite{quwaider2023shooter}. Several studies compared visual games with text-based games for personality assessment and found equivalent effectiveness, with players responding more positively to game-based methods than questionnaires \cite{harman2022illustrating}. Other research emphasized the role of games in reducing bias and minimizing incorrect responses, particularly through decision-based adventure games aligned with the Big Five framework \cite{harman2024advances}.

Further research investigated selection-oriented games and concluded that well-designed personality assessment games can be effectively used in high-stakes selection contexts. This study employed the HEXACO framework and reused a previously developed game \cite{barends2023construct}. Research conducted at Peter the Great University over five years applied gamification elements such as stations, quizzes, and Instagram competitions to analyze personality manifestations among first-year students \cite{yudina2021analysis}.

Escape-room-style games were also proposed, using the Big Five personality model to extract personality traits through diverse scenarios and mathematical modelling \cite{liapis2022modelling}. Other studies applied logistic regression and decision-tree models to assist employers in candidate selection, finding that cognitive predictions were more accurately extracted than personality traits through games \cite{wu2022individual}. Virtual reality environments were also introduced to simulate realistic social interactions for personality assessment \cite{miller2021virtual}. Additional research examined the relationship between personality and role selection in competitive games, including gender effects. Using data from 208 adult players, the study found no evidence to reject the null hypothesis regarding higher agreeableness among support-role players \cite{mayers2021relationship}.

Growing interest in gamified assessments stems from their ability to reduce cheating and elicit positive candidate reactions. Theoretical arguments suggest that gamification may discourage cheating by increasing cognitive load and obscuring which traits are being measured, thereby making it more difficult for candidates to intentionally present themselves in an overly favorable manner. One study attempted to investigate this by combining traditional and gamified personality assessments with explicit warnings about the detectability of dishonest responses, in order to examine the relationship between implicit game-based testing and response honesty \cite{yu2021impact}. Other small-scale studies confirmed stable correlations between game behavior and personality traits using the Big Five model \cite{palhano2020paki}.

Table 1 summarizes relevant studies published between 2019 and 2023, including the personality questionnaires employed, game types, and analytical methods. Studies with unspecified analytical methods or game types were excluded from the percentage calculations. The majority of the reviewed studies employed the Big Five personality framework, as illustrated in Figure 2. Furthermore, the figure shows that most studies favored self-developed games over commercial ones, and that statistical analysis was the dominant evaluation approach, followed by questionnaire-to-game mapping and machine-learning methods.

\begin{table}[ht]
\caption{Game-based personality assessment studies (2019-2023)}
\label{tab:game_personality}
\begin{center}
\begin{tabular}{|l|l|l|l|l|}
\hline
\rule[-1ex]{0pt}{3.5ex} Ref. & Year & Personality Model & Game Source & Analysis Method \\
\hline
\rule[-1ex]{0pt}{3.5ex} \cite{Elyasi2023mbtiC} & 2023 & MBTI & Self-Made & Relation (Statistical analysis) \\
\hline
\rule[-1ex]{0pt}{3.5ex} \cite{gao2023research} & 2023 & HEXACO & Commercial & Machine Learning \\
\hline
\rule[-1ex]{0pt}{3.5ex} \cite{quwaider2023shooter} & 2023 & Big Five & Self-Made & Machine Learning \\
\hline
\rule[-1ex]{0pt}{3.5ex} \cite{harman2022illustrating} & 2022 & Big Five & Self-Made & Questionnaire Items to Game Mechanics \\
\hline
\rule[-1ex]{0pt}{3.5ex} \cite{harman2024advances} & 2022 & Big Five & Self-Made & Relation (Statistical analysis) \\
\hline
\rule[-1ex]{0pt}{3.5ex} \cite{barends2023construct} & 2022 & HEXACO & Self-Made & Relation (Statistical analysis) \\
\hline
\rule[-1ex]{0pt}{3.5ex} \cite{yudina2021analysis} & 2022 & MBTI & Self-Made & Relation (Statistical analysis) \\
\hline
\rule[-1ex]{0pt}{3.5ex} \cite{waters2022exploring} & 2022 & Big Five & Self-Made & Relation (Statistical analysis) \\
\hline
\rule[-1ex]{0pt}{3.5ex} \cite{landers2022game} & 2022 & None & None & None \\
\hline
\rule[-1ex]{0pt}{3.5ex} \cite{liapis2022modelling} & 2022 & Big Five & Self-Made & Questionnaire Items to Game Mechanics \\
\hline
\rule[-1ex]{0pt}{3.5ex} \cite{wu2022individual} & 2021 & Big Five & Self-Made & Machine Learning \\
\hline
\rule[-1ex]{0pt}{3.5ex} \cite{afroza2021development} & 2021 & Big Five & Self-Made & Relation (Statistical analysis) \\
\hline
\rule[-1ex]{0pt}{3.5ex} \cite{miller2021virtual} & 2021 & None & Self-Made & None \\
\hline
\rule[-1ex]{0pt}{3.5ex} \cite{haizel2021personality} & 2021 & Big Five & Self-Made & Questionnaire Items to Game Mechanics \\
\hline
\rule[-1ex]{0pt}{3.5ex} \cite{mayers2021relationship} & 2021 & Big Five & Self-Made & Relation (Statistical analysis) \\
\hline
\rule[-1ex]{0pt}{3.5ex} \cite{yu2021impact} & 2021 & HEXACO & Self-Made & Relation (Statistical analysis) \\
\hline
\rule[-1ex]{0pt}{3.5ex} \cite{amin2025comparative} & 2021 & MBTI / Big Five & None & None \\
\hline
\rule[-1ex]{0pt}{3.5ex} \cite{palhano2020paki} & 2020 & Big Five & Self-Made & Relation (Statistical analysis) \\
\hline
\rule[-1ex]{0pt}{3.5ex} \cite{wirth2020assessing} & 2020 & HOSP & Self-Made & Relation (Statistical analysis) \\
\hline
\rule[-1ex]{0pt}{3.5ex} \cite{ponticorvo2020big} & 2020 & Big Five & Self-Made & Questionnaire Items to Game Mechanics \\
\hline
\rule[-1ex]{0pt}{3.5ex} \cite{raja2020gaming} & 2020 & Multi (DSM-5) & None & Relation (Statistical analysis) \\
\hline
\rule[-1ex]{0pt}{3.5ex} \cite{stuanescu2020game} & 2020 & None & None & None \\
\hline
\rule[-1ex]{0pt}{3.5ex} \cite{wang2019personality} & 2019 & Big Five & Commercial & Relation (Statistical analysis) \\
\hline
\rule[-1ex]{0pt}{3.5ex} \cite{olson2019predicting} & 2019 & Not Mentioned & Commercial & Relation (Statistical analysis) \\
\hline
\rule[-1ex]{0pt}{3.5ex} \cite{mccord2019game} & 2019 & Big Five & Self-Made & Relation (Statistical analysis) \\
\hline
\rule[-1ex]{0pt}{3.5ex} \cite{zulkifly2019personality} & 2019 & Big Five & Self-Made & Questionnaire Items to Game Mechanics \\
\hline
\end{tabular}
\end{center}
\end{table}

\begin{figure}[ht]
\centering
\includegraphics[width=0.8\linewidth]{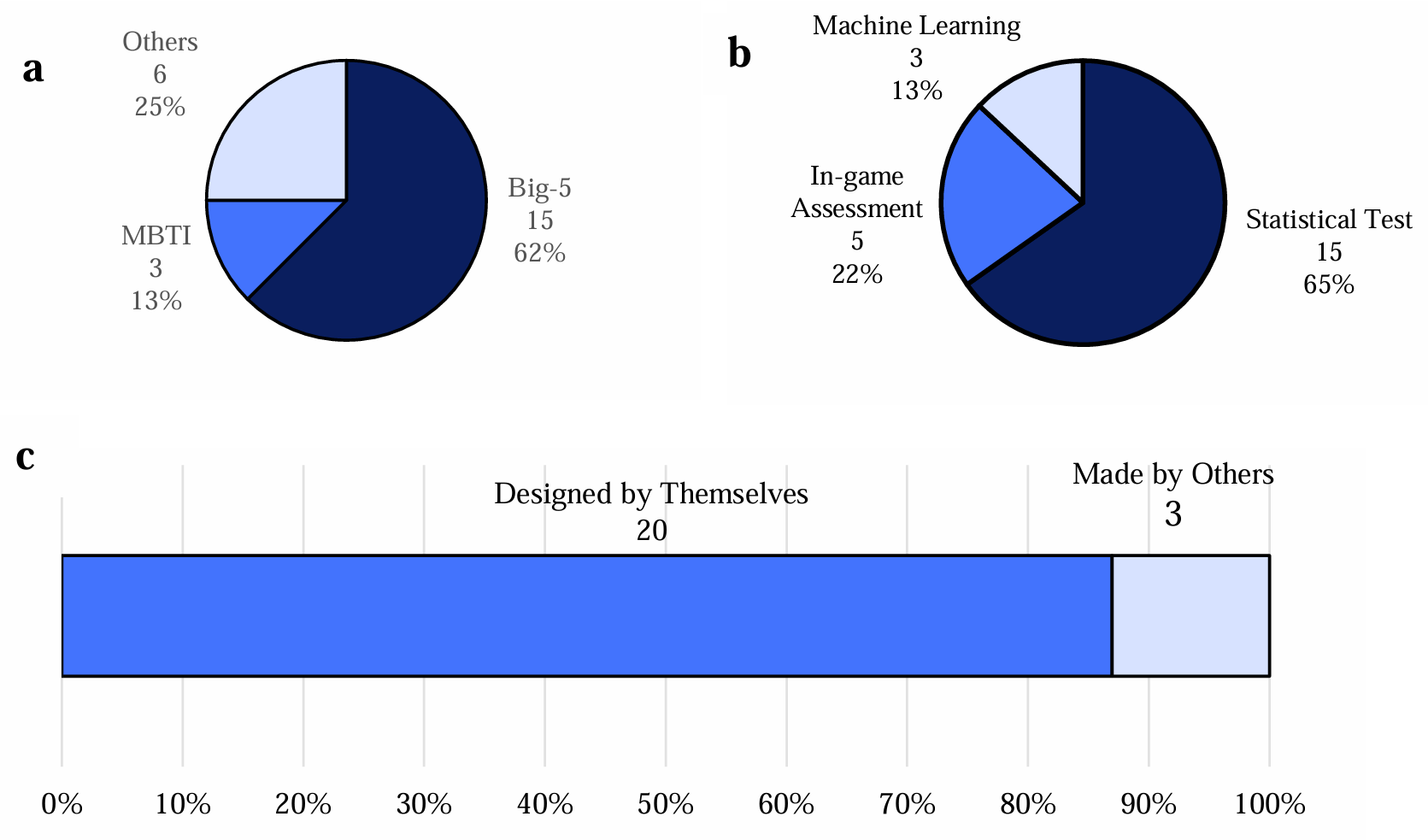}
\caption{Overview of prior game-based personality assessment studies included in the review (2019-2023). (a) Personality models used. (b) Analysis methods applied. (c) Source of games used in studies.
}
\label{fig:example}
\end{figure}

\section{Methodology}
\subsection{Identification of Developer-Relevant Traits}

The first stage of the proposed methodology focuses on identifying the personality characteristics and behavioral traits that are most strongly associated with software development. This stage is critical, as it provides the conceptual foundation for the game design and determines which behaviors are considered meaningful during analysis. To achieve this, a dual analytical strategy was employed, combining a systematic review of existing international studies with an independent empirical investigation conducted within Iran.

The review of prior studies revealed that research on the personality of software engineers often relies on student samples rather than practicing professionals \cite{varastehnezhad2025jungian, Elyasi2023mbtiJ, capretz2002software, barnes1975programmer}. While such studies provide useful insights, they may not accurately reflect the characteristics of individuals who actively work in software development roles. Moreover, cultural and social factors can significantly influence personality distributions and behavioral expression \cite{rehman2012mapping, rasch1992factors}. For these reasons, an empirical study was conducted on professional software engineers employed in Iranian companies, with the objective of validating and contextualizing findings reported in the literature \cite{Elyasi2023mbtiJ}.

Analysis of both the international studies and the Iranian dataset demonstrated that treating personality frameworks as rigid categorical systems, such as full Myers-Briggs personality types, offers limited explanatory power, particularly in limited samples \cite{Elyasi2023mbtiJ, capretz2002software}. Instead, analyzing the independent dimensions of personality provides a more robust and statistically meaningful representation \cite{Elyasi2023mbtiJ, barnes1975programmer, rehman2012mapping}. In particular, results derived from the MBTI consistently showed that the Thinking versus Feeling dimension exhibits the strongest and most stable association with software development roles \cite{varastehnezhad2025jungian, Elyasi2023mbtiJ, capretz2002software, barnes1975programmer}. Individuals who prefer analytical, logic-driven decision-making were found to be significantly overrepresented among developers, while individuals whose decisions are primarily guided by affective or interpersonal considerations were underrepresented. Thinking-type (T) individuals are more dominant and better suited for software development and therefore should be identified. Such individuals exhibit the following characteristics \cite{martin1997looking, schweda2005personality}:

•	Logical

•	Analytical

•	Impersonal in judgment (truth is prioritized over tact)

•	Task-oriented (rule-based) and systematic

•	Problem-solving oriented

•	Critical and skeptical thinking

•	Ideation, creativity, and innovation

•	Data-driven

Beyond this primary personality dimension, repeated patterns emerged across both the literature and the empirical data regarding behavioral traits relevant to software development. To ensure robustness, only traits mentioned more than three times across the reviewed sources were prioritised. Conceptually overlapping traits, especially those frequently linked to MBTI Thinking-related characteristics, were merged to reduce redundancy. As a result, redundant traits were removed from the final list of characteristics derived from the literature, yielding a more coherent and practically measurable set. These traits include systematic problem-solving, logical reasoning, attention to structure and detail, adaptability to new challenges, effective time management, continuous learning, and the ability to seek information or assistance when necessary \cite{rehman2012mapping, rasch1992factors, amin2020impact, vadlamani2020studying, rajakarthikeyanstudy, li2020distinguishes}. Motivation for achievement and a willingness to engage with complex or challenging tasks were also frequently reported \cite{celikten2017assigning, li2015makes, ahmed2015soft}. Collectively, these traits informed the selection of latent behavioral constructs that the serious game was designed to elicit and measure.

Overall, the results suggest that a suitable software developer is primarily characterised by the T trait in the MBTI framework, complemented by key behavioral attributes. In particular, the ability to seek information and assistance, manage time effectively, maintain motivation for success, adapt to changing conditions, and engage in continuous learning emerges as essential for effective and sustainable performance in software development.

\subsection{Technical Foundation}
The next stage of the methodology involved the design and implementation of a serious game to serve as the primary behavioral data collection instrument. The game was developed using the Unity game engine, selected for its flexibility, cross-platform capabilities, and suitability for instrumented research applications. All gameplay logic and interaction handling were implemented using the C\# programming language.

The Android platform was chosen for deployment to maximize accessibility and to ensure that participants from different backgrounds could easily install and use the application without requiring specialized hardware. To support detailed behavioral logging and subsequent analysis, a dedicated backend infrastructure was developed. This infrastructure was implemented using ASP.NET Core as a web-based application programming interface responsible for receiving gameplay events from client devices. Microsoft SQL Server was used as the primary data storage system, providing reliable and structured storage for high-volume event data. Entity Framework was employed to define data models and manage interactions between the application and the database.

Each gameplay action performed by a participant, including movements, selections, pauses, retries, and navigation events, was recorded as a structured event and transmitted to the backend server. Data transmission was performed securely, and all participants were informed about the nature of the collected data prior to participation. Explicit consent was obtained before any data collection took place. No personally identifiable information was stored as part of the gameplay dataset, ensuring participant privacy and compliance with ethical research standards.

\subsection{Design Strategy and Gameplay Structure}
The primary objective was not entertainment alone, but the creation of gameplay situations that require cognitive effort, strategic thinking, adaptation, and learning under constraints. Initially, two major game concepts were explored. The first was a narrative-driven card-based role-playing game inspired by decision-heavy mechanics, where players make choices that influence story progression. Although this design provided rich qualitative data, pilot testing revealed that narrative engagement and emotional involvement often overshadowed analytical decision-making. Player behavior was strongly influenced by personal preferences for story outcomes rather than problem-solving strategies, which reduced the interpretability of recorded actions in relation to software development traits. As a result, this prototype was excluded from the final design.

The second prototype was a tower-defense style game emphasizing strategic placement, resource management, and long-term planning. While this design demonstrated potential in measuring planning ability, it imposed a substantial learning curve. Many participants spent a significant portion of their session attempting to understand the game mechanics rather than engaging in meaningful problem-solving. Given the limited experimental timeframe and the need to minimize learning bias, this prototype was also discarded.

Based on these observations, the final design strategy adopted a modular, multi-stage approach composed of short, independent gameplay scenarios. Each stage was designed to be learned quickly while still requiring increasingly sophisticated reasoning as difficulty progressed. This approach enabled the assessment of multiple cognitive and behavioral traits within a single session, while minimizing fatigue, frustration, and confounding learning effects, as well as the impact of prior knowledge of similar game mechanics. Importantly, the independence of stages ensured that failure or difficulty in one task did not prevent participants from engaging with subsequent tasks.

The finalized game consists of several distinct stages, each designed to target specific behavioral constructs associated with software development. These stages span multiple genres, including constrained puzzle-solving, memory-based reasoning, real-time action under pressure, spatial manipulation, and abstract graph traversal. Although the mechanics differ across stages, they share a consistent design philosophy: each task imposes explicit constraints that force players to reason, plan, test hypotheses, and revise strategies based on feedback.

One category of stages focuses on constrained problem-solving tasks in which players must rearrange objects or entities within a limited space to reach a predefined goal state. These tasks are designed to measure systematic planning, foresight, and the ability to reason about sequences of actions before execution. As difficulty increases, the solution space expands combinatorially, making brute-force approaches inefficient and encouraging analytical reasoning. Player behavior in these stages provides insight into how individuals approach complex problems, whether they plan ahead or rely on trial and error, and how they respond when initial strategies fail.

Another category of stages centers on memory-based challenges. In these stages, players are briefly exposed to information that must later be recalled or inferred under time pressure. While short-term memory capacity plays a role, successful performance increasingly depends on the development of strategies, such as pattern recognition or systematic exploration. These stages were included not to assess memory in isolation, but to observe how participants adapt their approach when faced with incomplete information, a common characteristic of real-world software development tasks.

A further set of stages involves real-time survival scenarios in which players must continuously respond to dynamically changing conditions. These stages require rapid decision-making, prioritization of actions, and adaptation to unforeseen events. Unlike the puzzle-based stages, these scenarios offer greater freedom of movement and choice, enabling the observation of individual differences in risk tolerance, situational awareness, and learning from repeated failures. The richness of behavioral data generated in these stages allows for the extraction of multiple features related to adaptability and decision-making under pressure.

Additional stages emphasize spatial reasoning and abstract thinking. In spatial manipulation tasks, players must reposition obstacles or elements to guide an entity toward a target, requiring the anticipation of future states and the evaluation of alternative configurations. In graph traversal tasks, players are presented with interconnected structures governed by specific traversal rules and are required to find paths that satisfy global constraints. These stages are designed to elicit higher-level reasoning, mental simulation, and the ability to manage complex rule sets, all of which are central to software development activities such as algorithm design and debugging.

Across all stages, several cross-cutting design mechanisms were deliberately implemented to capture meaningful behavioral signals. Time limits were introduced to simulate real-world constraints and to observe how participants manage pressure. Optional tutorials were provided at the beginning of each stage, allowing participants to choose between guided instruction and exploratory learning. This choice itself serves as a behavioral indicator of learning preference and confidence. Additionally, the ability to pause and resume gameplay was included to capture strategic disengagement, reflection, and planning behavior.

A particularly important design element is the inclusion of voluntary surrender options. When players exceed the allotted time for a stage, they are given the choice to either continue attempting the task or to surrender and move on. This mechanism allows the measurement of persistence, frustration tolerance, and decision-making under perceived failure. Rather than treating surrender as a purely negative outcome, the methodology considers it a meaningful behavioral signal that can indicate adaptive self-regulation or avoidance, depending on context and frequency.

To further enrich the behavioral dataset, optional side challenges were incorporated into the game. These challenges, often based on mathematical or logical problems, allow players to earn resources that can be used to bypass particularly difficult stages. Engagement with these optional tasks provides insight into intrinsic motivation, willingness to invest additional effort, and attitudes toward learning and self-improvement. Importantly, players are not required to engage with these challenges, ensuring that participation reflects voluntary behavior rather than compliance.

The overall structure of the game was designed to balance control and freedom. While each stage imposes specific rules and constraints, players retain autonomy in how they approach and solve problems. This balance ensures that the collected data reflect genuine individual differences in cognitive style and behavioral strategy rather than uniform responses imposed by rigid mechanics. Further details regarding the games and their conditions are provided in Appendix A.

\subsection{Experimental Procedure and Data Collection}
Following the completion of game design and system implementation, the next stage of the methodology involved the execution of the experimental procedures and the systematic collection of data. This stage was designed to ensure that the collected behavioral and psychological data were both reliable and suitable for subsequent statistical and machine learning analysis. The experimental process was conducted in two main phases, consisting of a pilot study and a final large-scale experiment, each serving distinct methodological purposes.

The pilot study was conducted as an initial validation step to assess the usability of the game, the clarity of instructions, and the reliability of the event-logging mechanism. This preliminary experiment involved thirty-four participants drawn from a mixed population that included professional software engineers, graduate students in computer-related disciplines, and individuals with no formal background in computing. The inclusion of participants with diverse backgrounds allowed for the observation of a wide range of interaction patterns and helped identify potential design flaws that might disproportionately affect certain user groups. Gaming platform preferences, average gameplay duration, and the distribution of MBTI letters and paired types among participants in the first experiment can be seen in Figure 3.

\begin{figure}[ht]
\centering
\includegraphics[width=0.9\linewidth]{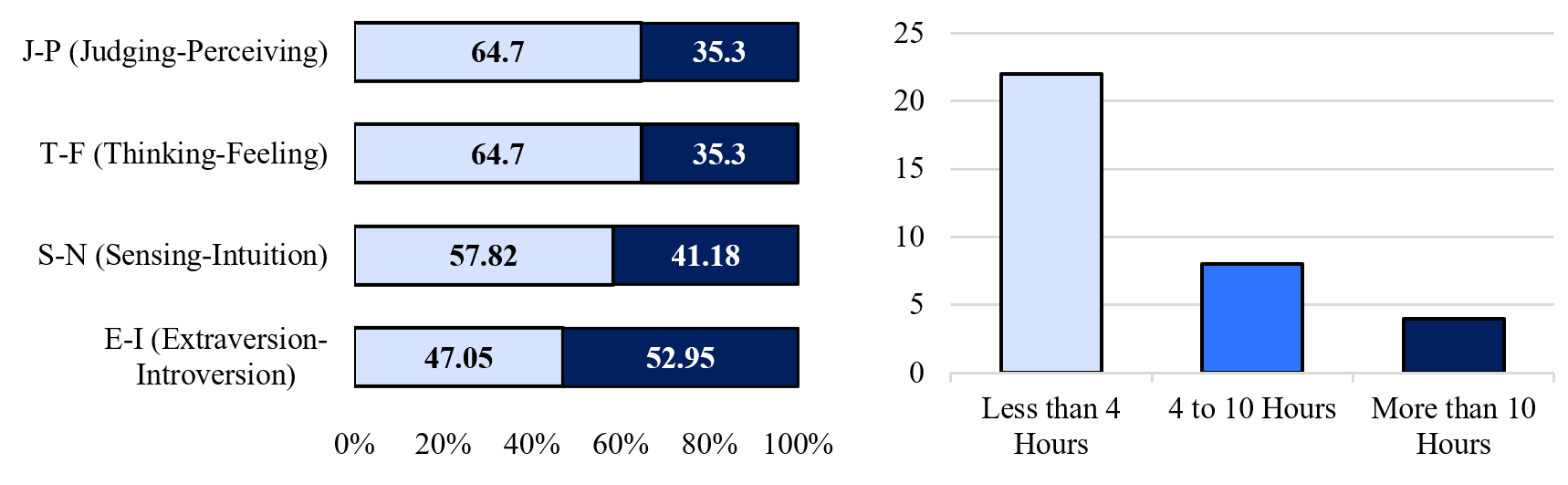}
\caption{
The distribution of MBTI letters, and the average time spent playing games 
}
\label{fig:example}
\end{figure}

During the pilot study, participants completed the full gameplay sequence and the associated personality questionnaire. Observations and informal feedback indicated that, while the overall structure of the game was understandable, certain stages required clearer explanations to reduce initial confusion. In response to these findings, tutorial content was refined, visual cues were enhanced, and difficulty progression was adjusted to ensure that early-stage challenges functioned as learning opportunities rather than barriers. Additionally, the granularity of behavioral logging was improved to ensure that subtle actions, such as hesitation before movement or repeated exploratory actions, were consistently captured. The pilot study also provided initial evidence that the game was capable of eliciting meaningful behavioral variation across participants, thereby supporting its suitability for the main experiment.

After incorporating refinements informed by the pilot phase, the final experimental study was conducted. A total of one hundred and fifty-three participants were recruited for this phase. Participants were drawn from secondary schools, universities, and professional environments, ensuring diversity in age, educational background, and familiarity with computing concepts. To ensure the validity of personality assessment instruments, participation was limited to individuals aged fourteen years or older. Ethical approval was obtained, and informed consent was secured from all participants prior to data collection. Moreover, they had the option to choose whether to send their anonymized data to our servers or to withhold it at the final stage of the games. If they chose to send it, they received a code to submit alongside their questionnaire, allowing the two to be linked. Each participant completed three components as part of the experiment.

•	They responded to a standardized Myers-Briggs personality questionnaire, which provided independent personality dimension scores used in the subsequent analysis. Participants were divided into two groups: one group completed the questionnaire before playing the game, while the other completed it after the game, in order to reduce potential mutual influence between the two.

•	Participants completed a demographic and self-assessment survey capturing information such as educational level, professional experience, interest in software development, and self-perceived competence in technical tasks.

•	Participants engaged with the full sequence of game stages, during which all interactions were recorded by the backend system.

Following data collection, the dataset was examined for completeness and quality. Records containing incomplete questionnaire responses, interrupted gameplay sessions, or corrupted event logs were excluded from further analysis. After this cleaning process, one hundred and thirty-two participant records remained and were deemed valid for subsequent modeling. This final dataset formed the empirical basis for all remaining stages of the methodology.

The gameplay sessions generated a high-dimensional behavioral dataset capturing fine-grained aspects of participant interaction. Initially, one hundred and thirty-two distinct gameplay-derived variables were extracted. These variables represented a broad range of behaviors, including but not limited to the frequency of actions, number of retries, error counts, time spent per stage, use of tutorials, frequency of pauses, instances of surrender, navigation through menus, and engagement with optional side challenges. Each variable was designed to correspond, either directly or indirectly, to one or more of the developer-relevant traits identified earlier in the study. List of these variables can be seen in the appendix B.

To prepare the dataset for analysis, an extensive preprocessing phase was conducted. Variables with insufficient data coverage, such as those corresponding to rarely triggered events, were removed to prevent noise and instability in subsequent models. Features exhibiting near-zero variance across participants were also excluded, as they provide little discriminative information. In addition, correlation analysis was performed to identify highly correlated variables that could introduce redundancy and negatively impact model performance. In cases where strong correlations were observed, the more interpretable or theoretically relevant variable was retained. After completing these preprocessing steps, approximately one hundred and twenty behavioral variables remained.

In parallel with behavioral data preprocessing, a labeling strategy was implemented to enable supervised learning. A subset of participants was labeled based on confirmed expertise in software development, including verified academic or professional involvement in the field. This verification was conducted using self-reported survey responses in combination with expert judgment. Approximately forty participants satisfied these criteria and were assigned binary labels indicating suitability or non-suitability for software development. These labeled instances served as ground truth for model training and evaluation. The remaining participants were treated as unlabeled cases whose suitability was to be inferred by the predictive models.

This labeling approach reflects the practical constraints of real-world career guidance scenarios, in which definitive labels are often scarce and expensive to obtain. By relying on a limited but carefully curated set of labeled examples, the methodology aims to balance realism with analytical rigor. The presence of unlabeled data also motivated the use of robust modeling techniques capable of generalizing from limited supervision.

At the conclusion of this stage, the dataset consisted of three tightly integrated components: independent personality dimensions derived from the MBTI questionnaire, demographic and self-assessment variables, and a high-dimensional behavioral representation extracted from gameplay. Together, these components provided a rich and multifaceted foundation for the machine learning analyses described in the subsequent section. Demographic information and MBTI personality profiles of players in the second experiment can be seen in Figure 4.

\begin{figure}[ht]
\centering
\includegraphics[width=0.8\linewidth]{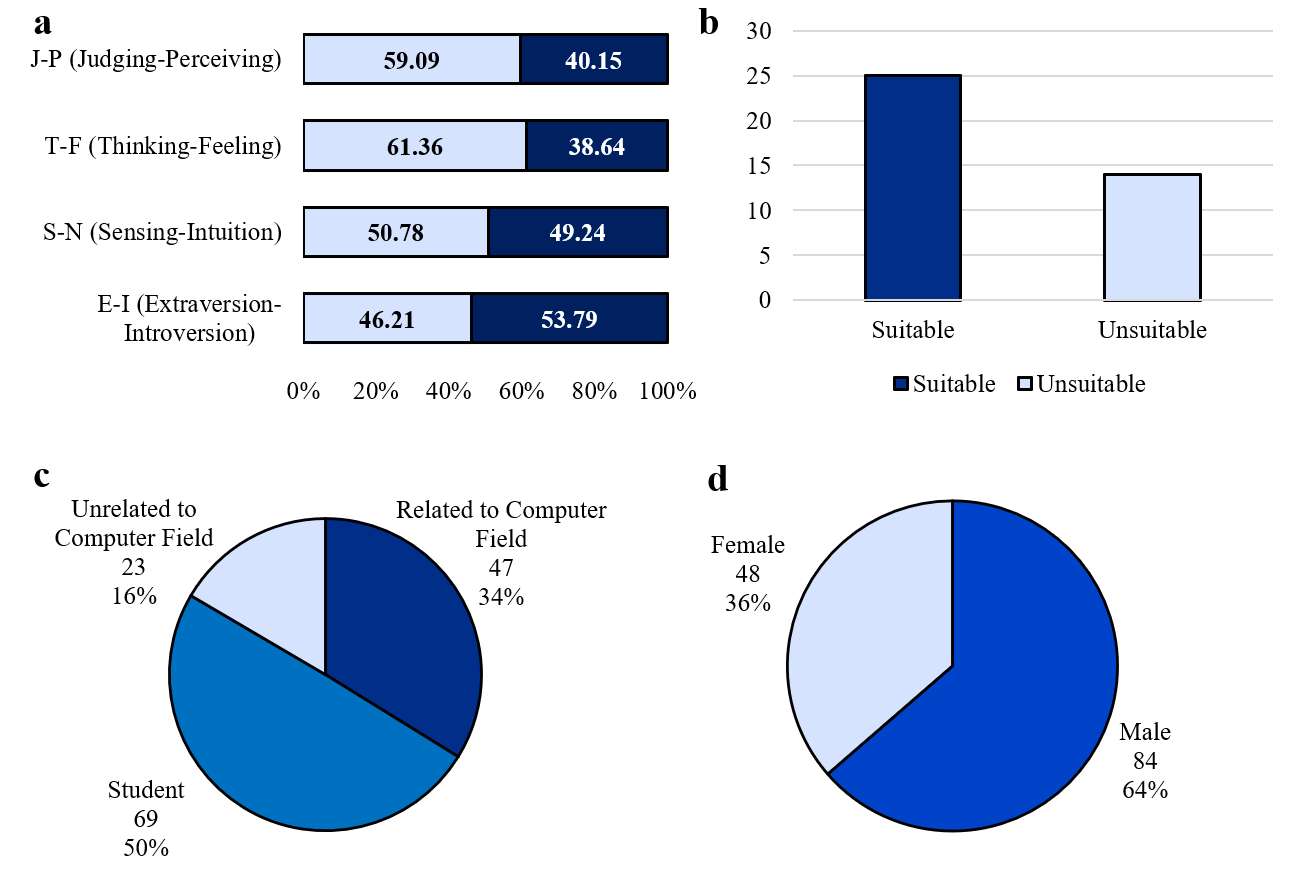}
\caption{
MBTI personality dimensions and demographic characteristics of participants in the second experiment. (a) Percentage distribution of the four MBTI dichotomies. (b) Number of participants classified as suitable or unsuitable for computer-industry roles based on evidence and expert judgment. (c) Participant background distribution. (d) Gender distribution of participants.}
\label{fig:example}
\end{figure}

\subsection{Machine Learning Framework and Modeling Strategy }
Following the completion of data collection, preprocessing, and labeling, the methodology proceeded to the analytical and modeling stage. The objective of this stage was to construct predictive models capable of inferring suitability for software development based on the collected personality data (with an emphasis on T-type personality traits), additional questionnaire responses, and a small, verified subset of suitable and unsuitable cases. The additional questions were derived from characteristics frequently associated in the literature with T-type traits within the MBTI framework; however, characteristics that directly overlapped with the T dimension itself were explicitly excluded to avoid duplicate evaluation, as the T personality dimension was already used as an independent variable.

To achieve this objective in a structured and interpretable manner, the machine-learning framework was divided into two sequential phases. The first phase focused on completing the labeled dataset using personality-based information, while the second phase concentrated on re-predicting suitability across the entire dataset using gameplay-derived behavioral features alone.

The first modeling phase addressed the problem of dataset completion. Given that only a subset of participants had clearly defined labels indicating suitability for software development, this phase aimed to predict labels for the remaining participants using psychological and self-reported information. The task was formulated as a binary classification problem, in which each participant was assigned to either a suitable or non-suitable class. The input features for this phase consisted primarily of independent personality dimensions derived from the Myers-Briggs questionnaire, together with a limited number of self-assessment variables reflecting learning behavior, adaptability, time management, and information-seeking tendencies.

Several supervised learning algorithms were evaluated during this phase. Logistic regression was selected as a baseline model due to its interpretability and robustness when working with small datasets. In addition, ensemble-based methods such as random forest classifiers were employed to capture potential nonlinear relationships between personality dimensions and suitability outcomes. Support vector machines were also explored, as they are well suited for binary classification problems with limited samples and can effectively handle high-dimensional feature spaces. Gradient boosting methods were included to assess whether sequential ensemble learning could improve predictive performance by iteratively correcting classification errors.

Given the relatively small number of labeled instances, controlling overfitting was a primary concern throughout this phase. To address this issue, cross-validation was employed during model training and evaluation. The dataset was partitioned into multiple folds, and each model was trained and tested across different partitions to ensure that performance estimates were stable and not dependent on a particular split of the data. Accuracy was selected as the primary evaluation metric in this phase, as both false positive and false negative predictions have practical implications in career recommendation contexts. Models that demonstrated stable performance across folds were retained for use in labeling the remaining participants.

Upon completion of the first phase, the dataset contained a completed set of labels representing inferred suitability for software development. This enriched dataset served as the foundation for the second and more critical modeling phase, which aimed to predict suitability using gameplay-derived behavioral features exclusively. This phase constitutes the core methodological contribution of the study, as it evaluates whether implicit behavioral signals captured during gameplay can reliably replace or supplement traditional personality assessments.

Before model training, extensive preprocessing and feature engineering were performed on the behavioral dataset. Although earlier preprocessing steps had already removed low-variance and redundant features, the high dimensionality of the remaining behavioral variables posed additional challenges. Feature scales varied substantially, reflecting differences in action counts, time measurements, and binary indicators. To mitigate the influence of scale disparities on model performance, normalization techniques were applied to rescale features into comparable ranges. This step was particularly important for distance-based algorithms and gradient-based optimization methods.

Feature selection and dimensionality reduction played a central role in this phase of the methodology. Given the large number of behavioral variables relative to the number of labeled instances, reducing dimensionality was essential to improve model generalization and interpretability. Multiple feature selection strategies were explored. Univariate statistical methods were used to evaluate the individual relationship between each feature and the target label, allowing the identification of variables with strong discriminatory power. Recursive feature elimination was applied in conjunction with supervised learning models to iteratively remove less informative features based on model performance. In parallel, tree-based feature importance measures derived from ensemble models were used to identify variables that contributed most strongly to prediction accuracy.

In addition to feature selection, dimensionality reduction techniques were investigated to capture latent behavioral patterns in a compact form. Linear methods were employed to project the original feature space into lower-dimensional representations that preserve class separability, while nonlinear methods were explored to capture complex relationships among features. These reduced representations were then used as inputs to the classification models, allowing the evaluation of whether abstract behavioral embeddings could improve predictive performance relative to raw or selected features.

Model training in this phase employed stratified k-fold cross-validation to ensure that class distributions were preserved across folds. This approach was particularly important given the presence of class imbalance, as the number of participants labeled as non-suitable was smaller than the number labeled as suitable. To further address this imbalance, experiments were conducted both with and without random oversampling of the minority class. Oversampling was applied only to the training folds to prevent information leakage and to maintain the integrity of evaluation results.

Multiple classification models were trained and evaluated across different feature subsets, dimensionality reduction strategies, and sampling configurations. Performance was assessed using accuracy, precision, recall, and F1-score, providing a balanced view of model effectiveness across different error types. In addition to quantitative metrics, qualitative analysis was performed to assess model stability, sensitivity to feature selection, and consistency across validation folds. Particular attention was given to identifying signs of overfitting, such as large discrepancies between training and validation performance or excessive dependence on small subsets of features.

The outcome of this modeling phase was the identification of one or more models that demonstrated robust and stable performance in predicting software development suitability based solely on gameplay behavior. These models form the basis for the proposed game-based career recommendation framework and provide empirical support for the hypothesis that implicit behavioral assessment can serve as a viable alternative to explicit personality testing.

\subsection{Model Evaluation, Selection, and Robustness Analysis}

Following model development and training, the final stage focused on evaluation, selection, and robustness analysis. This stage aimed not only to identify high-performing models, but also to ensure stability, generalization ability, and methodological soundness. Given the system’s intended use in career guidance, both quantitative performance and qualitative model behavior were considered. Model evaluation employed multiple complementary metrics. Accuracy was used as an overall indicator of correct classification but, due to potential class imbalance, was supplemented with precision, recall, and F1-score. Precision assessed the reliability of positive suitability predictions, recall measured the ability to identify genuinely suitable individuals, and the F1-score balanced these two aspects into a single interpretable measure.

\section{Results}

\subsection{Completion of Suitability Labels Using Personality-Based Features}
The first modeling phase aimed to complete missing suitability labels using personality dimensions derived from the Myers-Briggs questionnaire together with supplementary self-assessment features. A subset of 39 participants with confirmed suitability for software development was used as ground truth. The remaining records were treated as unlabeled cases. The task was formulated as a binary classification problem and evaluated using k-fold cross-validation. The comparative performance of the evaluated machine-learning algorithms is reported in Table 2.

\begin{table}[ht]
\caption{Model performance for personality-based suitability labeling}
\label{tab:model_performance}
\begin{center}
\begin{tabular}{|l|l|l|l|l|}
\hline
\rule[-1ex]{0pt}{3.5ex} Algorithm & F1 Score & Accuracy & Precision & Recall \\
\hline
\rule[-1ex]{0pt}{3.5ex} Random Forest & 0.76 & 0.80 & 0.76 & 0.76 \\
\hline
\rule[-1ex]{0pt}{3.5ex} MLP-64h & 0.76 & 0.80 & 0.76 & 0.76 \\
\hline
\rule[-1ex]{0pt}{3.5ex} MLP-22-63h & 0.76 & 0.80 & 0.76 & 0.76 \\
\hline
\rule[-1ex]{0pt}{3.5ex} SVM & 0.67 & 0.75 & 0.67 & 0.67 \\
\hline
\rule[-1ex]{0pt}{3.5ex} Logistic Regression & 0.41 & 0.70 & 0.35 & 0.50 \\
\hline
\rule[-1ex]{0pt}{3.5ex} GBM & 0.33 & 0.50 & 0.33 & 0.33 \\
\hline
\end{tabular}
\end{center}
\end{table}

As shown in Table 2, Random Forest and MLP achieved the highest accuracy, equal to 0.80, with balanced precision and recall values (both equal to 0.76). These models exhibited stable performance across validation folds. SVM and Logistic Regression achieved lower accuracy, while GBM showed the weakest performance.

Given that this phase was intended to complete dataset labels rather than optimize a final predictive model, accuracy was considered the most important metric. Based on this criterion, Random Forest was selected to infer suitability labels for the remaining participants.

To evaluate whether reducing the number of input variables could improve performance, correlation analysis was conducted between the six personality-based features and the suitability label. The resulting correlation matrix is illustrated in Figure 5. Reducing the feature set to the four most correlated variables did not lead to any improvement in accuracy or model stability. Inspection of training and validation metrics confirmed that the selected models did not exhibit overfitting.

\begin{figure}[ht]
\centering
\includegraphics[width=0.7\linewidth]{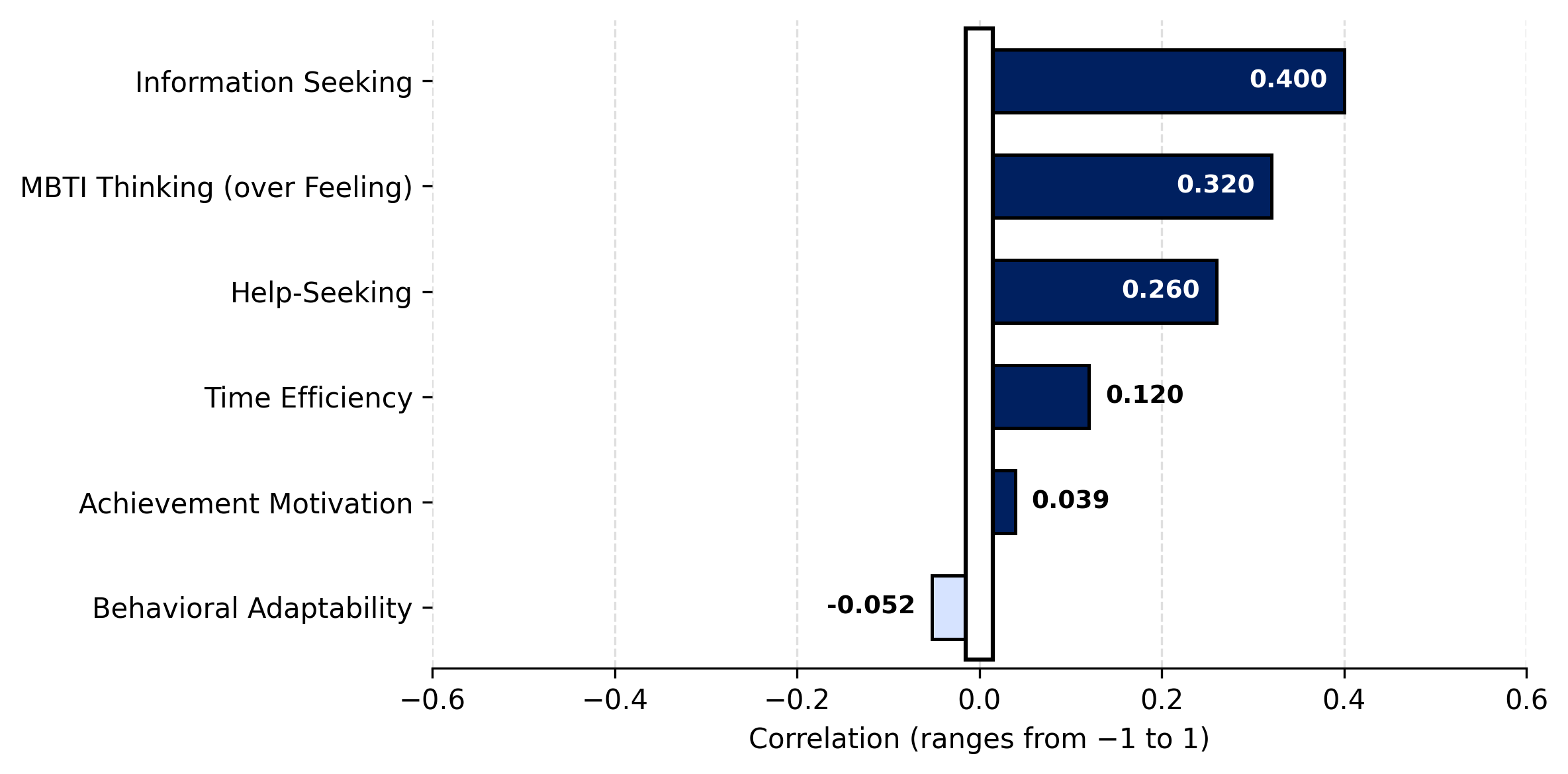}
\caption{
Correlation between personality features and suitability labels
}
\label{fig:example}
\end{figure}

\subsection{Prediction of Software-Development Suitability Using Gameplay Data}

In the second modeling phase, software-development suitability was predicted exclusively from gameplay-derived behavioral features, using labels generated in the first phase and excluding questionnaire-based or personality data. This phase assessed whether implicit behavioral signals captured during gameplay are sufficient for reliable prediction. To address this question, two complementary modeling strategies were examined: feature selection and feature reduction. Feature-selection results, summarized in Table 3, show that Random Forest models consistently achieved strong performance, with precision up to 0.94 and accuracy up to 0.88. Oversampling improved recall in some cases but did not consistently enhance overall accuracy. The impact of oversampling and algorithm-wise performance is illustrated in Figure 6.

\begin{table}[ht]
\caption{Gameplay-based suitability prediction using feature selection}
\label{tab:feature_selection}
\begin{center}

\setlength{\tabcolsep}{3pt} 

\begin{tabular}{|l|l|l|l|l|l|l|}
\hline
\rule[-1ex]{0pt}{3.5ex} Algorithm & Feature Selection & \#Features & Precision & Accuracy & Recall & F1 Score \\
\hline
\rule[-1ex]{0pt}{3.5ex} Random Forest & Based on Correlation & -- & 0.94 & 0.88 & 0.67 & 0.72 \\
\hline
\rule[-1ex]{0pt}{3.5ex} Random Forest & Univariate Feature Selection & 10 & 0.94 & 0.88 & 0.67 & 0.72 \\
\hline
\rule[-1ex]{0pt}{3.5ex} Random Forest & Random Forest & 5 & 0.94 & 0.88 & 0.67 & 0.72 \\
\hline
\rule[-1ex]{0pt}{3.5ex} Random Forest & Random Forest & 15 & 0.94 & 0.88 & 0.67 & 0.72 \\
\hline
\rule[-1ex]{0pt}{3.5ex} MLP & Recursive Feature Elimination & 15 & 0.94 & 0.88 & 0.67 & 0.72 \\
\hline
\rule[-1ex]{0pt}{3.5ex} Logistic Regression & Univariate Feature Selection & 10 & 0.92 & 0.85 & 0.58 & 0.60 \\
\hline
\rule[-1ex]{0pt}{3.5ex} Random Forest Oversample & Recursive Feature Elimination & 5 & 0.92 & 0.85 & 0.58 & 0.60 \\
\hline
\rule[-1ex]{0pt}{3.5ex} GBM Oversample & Recursive Feature Elimination & 10 & 0.80 & 0.89 & 0.80 & 0.80 \\
\hline
\rule[-1ex]{0pt}{3.5ex} Logistic Regression & Random Forest & 5 & 0.77 & 0.85 & 0.65 & 0.68 \\
\hline
\rule[-1ex]{0pt}{3.5ex} Random Forest & Random Forest & 10 & 0.77 & 0.85 & 0.65 & 0.68 \\
\hline
\rule[-1ex]{0pt}{3.5ex} Random Forest Oversample & Random Forest & 15 & 0.77 & 0.85 & 0.65 & 0.68 \\
\hline
\rule[-1ex]{0pt}{3.5ex} MLP Oversample & Recursive Feature Elimination & 15 & 0.77 & 0.85 & 0.65 & 0.68 \\
\hline
\end{tabular}

\end{center}
\end{table}

Feature-reduction results are reported in Table 4, where LDA, PCA, t-SNE, and UMAP were tested with multiple classifiers, both with and without oversampling. The best overall performance was obtained using LDA combined with an MLP classifier and oversampling, achieving 0.97 precision and 0.94 accuracy. Nonlinear reduction methods (t-SNE and UMAP) yielded lower accuracy and less stable results. The effects of oversampling and algorithm comparisons are shown in Figures 6.

\begin{table}[ht]
\caption{Gameplay-based suitability prediction using feature reduction}
\label{tab:feature_reduction}
\begin{center}
\begin{tabular}{|l|l|l|l|l|l|}
\hline
\rule[-1ex]{0pt}{3.5ex} Algorithm & Feature Reduction & Precision & Accuracy & Recall & F1 Score \\
\hline
\rule[-1ex]{0pt}{3.5ex} MLP Oversample & LDA & 0.97 & 0.94 & 0.84 & 0.88 \\
\hline
\rule[-1ex]{0pt}{3.5ex} Logistic Regression & LDA & 0.95 & 0.91 & 0.75 & 0.81 \\
\hline
\rule[-1ex]{0pt}{3.5ex} Random Forest & LDA & 0.95 & 0.91 & 0.75 & 0.81 \\
\hline
\rule[-1ex]{0pt}{3.5ex} GBM & LDA & 0.95 & 0.91 & 0.75 & 0.81 \\
\hline
\rule[-1ex]{0pt}{3.5ex} MLP & LDA & 0.95 & 0.91 & 0.75 & 0.81 \\
\hline
\rule[-1ex]{0pt}{3.5ex} Random Forest Oversample & LDA & 0.95 & 0.91 & 0.75 & 0.81 \\
\hline
\rule[-1ex]{0pt}{3.5ex} GBM Oversample & LDA & 0.95 & 0.91 & 0.75 & 0.81 \\
\hline
\rule[-1ex]{0pt}{3.5ex} GBM Oversample & PCA & 0.82 & 0.88 & 0.73 & 0.76 \\
\hline
\rule[-1ex]{0pt}{3.5ex} Logistic Regression Oversample & LDA & 0.80 & 0.88 & 0.80 & 0.80 \\
\hline
\rule[-1ex]{0pt}{3.5ex} MLP & T-SNE & 0.58 & 0.79 & 0.55 & 0.55 \\
\hline
\rule[-1ex]{0pt}{3.5ex} MLP Oversample & T-SNE & 0.54 & 0.78 & 0.53 & 0.53 \\
\hline
\rule[-1ex]{0pt}{3.5ex} MLP Oversample & UMAP & 0.53 & 0.67 & 0.54 & 0.53 \\
\hline
\end{tabular}
\end{center}
\end{table}

\begin{figure}[ht]
\centering
\includegraphics[width=0.7\linewidth]{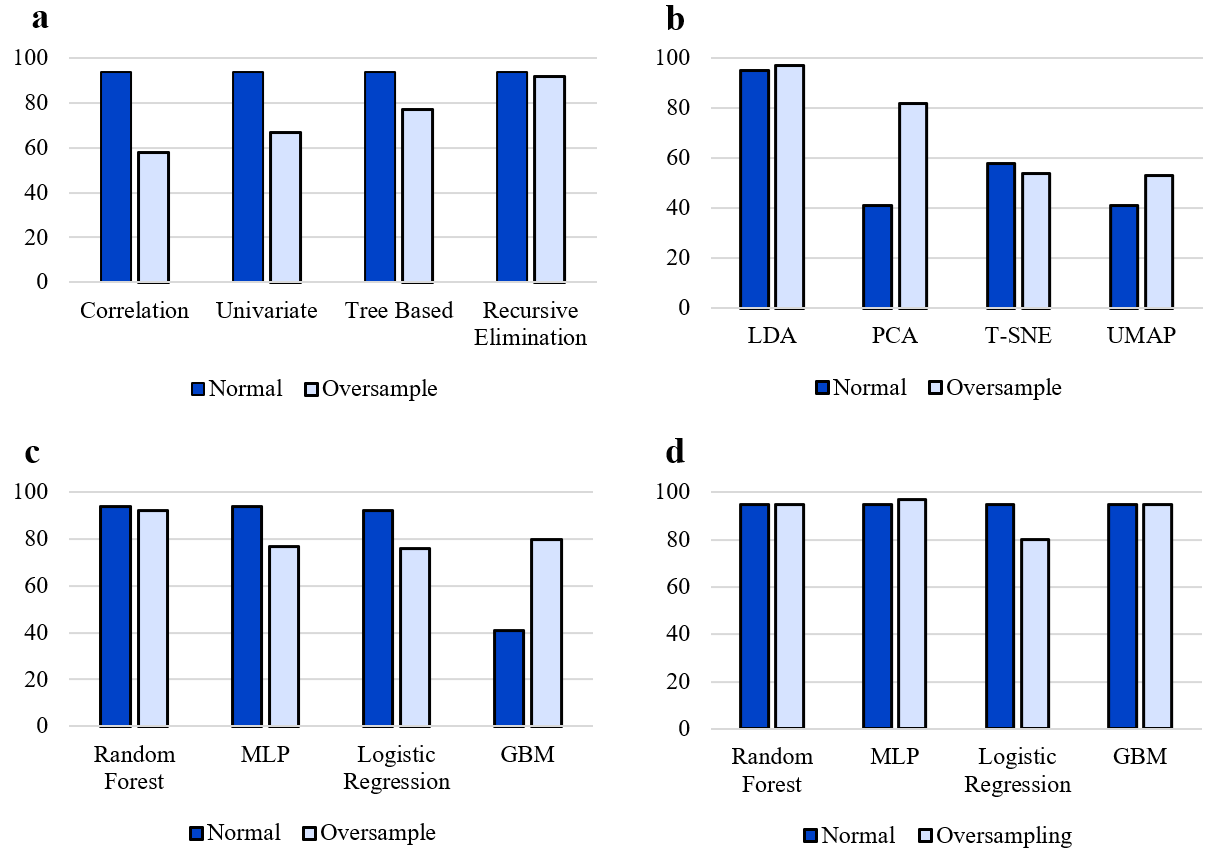}
\caption{
Comparison of feature-selection and feature-reduction results (a) Feature-selection methods, where for each method the best-performing classifier is reported under normal training and with oversampling. (b) Feature-reduction techniques, showing the highest-performing classifier associated with each reduction method, compared between normal training and oversampling. (c) Algorithm-wise comparison for feature-selection approaches, illustrating the performance of each classifier aggregated over the selected feature subsets, under normal and oversampled settings. (d) Algorithm-wise comparison for feature-reduction approaches, showing classifier performance when trained on reduced feature representations, with and without oversampling.
}
\label{fig:example}
\end{figure}

Overall, the two modeling phases demonstrate that personality-based features can complete missing suitability labels with 80\% accuracy (Table 2), while gameplay-derived behavioral features alone can predict software-development suitability with up to 97\% precision and 94\% accuracy (Table 4), providing strong quantitative evidence for the effectiveness of serious-game behavioral data in suitability prediction.

\section{Discussion}

Prior to building the final gameplay-based prediction models, exploratory correlation analyses were conducted to investigate relationships between gameplay-derived behavioral features and four previously identified developer-related traits. For interpretability, only correlations with an absolute value exceeding 0.25 were retained. As shown in Figure 7, stronger search-related traits were associated with more careful review of tutorials before gameplay.

Thinking-oriented traits were linked to fewer recorded events, fewer guesses in memory-based stages, and higher success rates in puzzle-oriented games, but to weaker performance in fast-paced shooter-style stages, reflected by higher event counts. Participants more willing to seek help engaged more extensively with instructional content prior to gameplay. Stronger time-management ability corresponded to fewer overall gameplay events and improved performance in the shooter-style game.

Finally, correlations between inferred software-developer suitability labels and gameplay behaviors (Figure 7) indicate that suitable participants achieved more wins in puzzle-based games, completed more side challenges, navigated menus more frequently, and exhibited fewer pauses, retries, and surrender actions.

\begin{figure}[ht]
\centering
\includegraphics[width=0.6\linewidth]{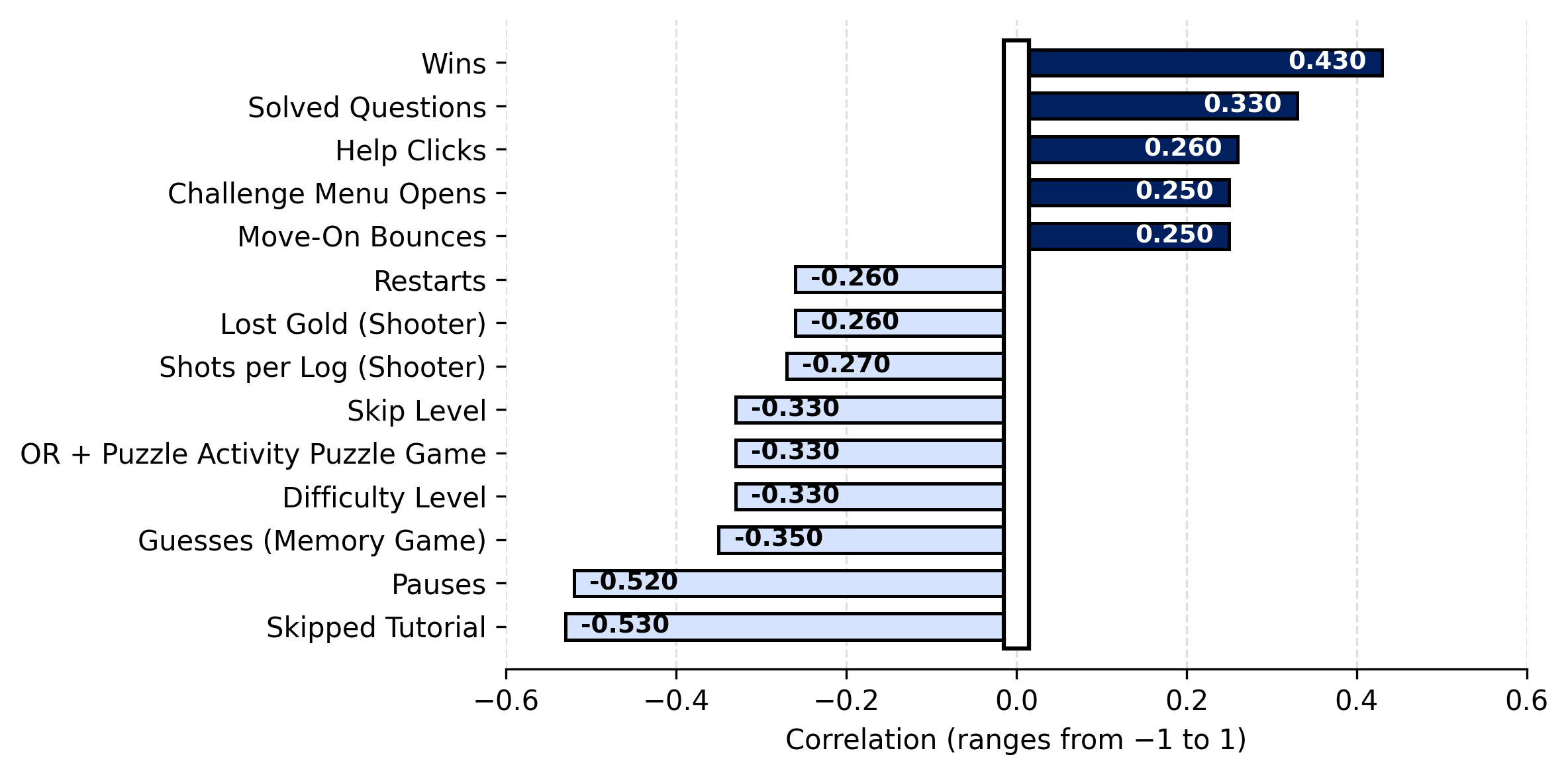}
\caption{
Correlation between gameplay behaviors and suitability labels
}
\label{fig:example}
\end{figure}

The strong performance of the Multilayer Perceptron (MLP) model, particularly when combined with Linear Discriminant Analysis (LDA) and oversampling, is not incidental, but rather reflects a good match between the model architecture and the structure of the data.

First, the gameplay dataset is high-dimensional and continuous, consisting of numerous interdependent behavioral indicators such as action frequencies, time measures, and interaction patterns. These features do not follow simple linear relationships with the suitability label. MLPs are well suited to this setting because they can model nonlinear combinations of features and capture interactions that are difficult to express using linear classifiers.

Second, the use of LDA as a feature-reduction step plays a critical role in stabilizing the MLP. LDA projects the original behavioral feature space into a lower-dimensional representation that explicitly maximizes class separability. This transformation reduces noise and redundancy while preserving the most discriminative behavioral information. As a result, the MLP operates on a compact, well-structured input space, which mitigates overfitting and improves generalization, an important consideration given the dataset size.

Third, comparison with other models further supports the rationality of using MLP. Random Forest models performed well and achieved competitive accuracy and precision, indicating that ensemble-based approaches can also capture relevant behavioral patterns. However, Random Forests rely on axis-aligned splits and may struggle to represent smooth decision boundaries in continuous spaces. In contrast, MLPs can learn smooth, distributed representations, which is advantageous when behavioral traits emerge from combinations of multiple weak signals rather than a small number of dominant features.

Finally, the inferior performance of nonlinear dimensionality-reduction techniques such as t-SNE and UMAP reinforces the appropriateness of the selected approach. While these methods are effective for visualization, they do not preserve class-discriminative structure in a supervised learning context. The fact that LDA-based representations consistently outperformed these alternatives indicates that the discriminative information in the gameplay data is best captured through supervised linear projections followed by nonlinear classification, precisely the configuration implemented in the LDA + MLP pipeline.

The discussion of both behavioral and modeling results highlights a key implication of this research: game-based assessment enables a shift from self-reported traits to observed competence. By capturing how individuals plan, adapt, persist, and prioritize under constraints, serious games provide access to decision-level behavior that is difficult to fake or consciously manipulate.

From a practical perspective, this suggests that game-based systems could be used as a scalable, engaging, and less biased tool for early-stage career guidance or candidate screening. Importantly, the results indicate that such systems do not need to rely on explicit personality questionnaires, which are subject to self-report bias and fatigue, but can instead infer relevant traits implicitly through gameplay.

\section{Limitations and Future Work}

Despite the encouraging results presented in this study, several limitations must be acknowledged when interpreting the findings, and these limitations also point toward important directions for future research.

A primary limitation concerns the relatively small subset of participants with confidently verified suitability labels for software development. Although the full dataset consisted of 132 valid participants, only 39 individuals were labeled as suitable or non-suitable based on confirmed academic or professional background. These labels formed the foundation of the supervised learning process. While cross-validation and multiple modeling strategies were employed to reduce overfitting, the limited size of the labeled subset may still introduce uncertainty in the learned decision boundaries. Future work should aim to expand the pool of verified labels, particularly through collaboration with industry partners or longitudinal tracking of career outcomes.

Suitability labels were derived from a combination of self-reported background information and expert judgment, rather than direct measurement of long-term job performance or career success. Although this approach reflects realistic constraints in early-stage career guidance research, it introduces potential labeling noise. Some participants classified as suitable may not ultimately succeed in professional software-development roles, and vice versa. Future studies could strengthen validity by incorporating objective performance indicators, such as academic achievement, coding assessments, internship evaluations, or workplace performance reviews collected over time. Moreover, future work could explore the use of regression models instead of classification approaches to estimate the degree of suitability for a given job. Such models could provide continuous suitability scores, offering more nuanced insights into how suitable an individual is and identifying specific skill areas where improvement could enhance their job readiness.

Additionally, extending gameplay to evaluate participants at different times of the day and across multiple days could further improve the robustness of the results. Human performance and behavior are influenced by fluctuating factors such as mood, fatigue, and cognitive state; therefore, collecting data under varied temporal conditions, whether through games or questionnaires, can provide a more reliable and representative assessment of individual characteristics.

Although each game stage was designed to target specific cognitive or behavioral constructs, it is not possible to guarantee that every gameplay mechanic exclusively measured its intended trait. For example, performance in puzzle-based stages may reflect not only planning ability but also prior exposure to similar games or abstract reasoning skills. Likewise, time-pressured stages may simultaneously engage stress tolerance, motor skills, and familiarity with action games. While the multi-stage design helps mitigate this issue by triangulating behavior across tasks, future work could incorporate formal construct-validation studies or controlled ablation experiments to better isolate individual trait contributions.

The strongest predictive performance was achieved using a combination of Linear Discriminant Analysis and Multilayer Perceptron models. While effective, neural models provide limited transparency regarding which specific behavioral patterns drive individual predictions. For deployment in high-stakes contexts such as career guidance or personnel screening, explainability is crucial. Future research should integrate explainable machine-learning techniques, such as SHAP or feature-attribution analyses, to improve transparency and user trust.

The proposed framework was specifically designed to assess suitability for software development. Although many of the captured behavioral traits, such as planning, adaptability, and problem-solving, are relevant to other technical professions, the game mechanics and modeling pipeline were optimized for this domain. Extending the framework to other career paths would require redesigning certain stages and recalibrating trait mappings. This represents a promising avenue for future expansion. Finally, these tools require long-term evaluation under controlled or clinical settings to establish their reliability and applicability in real-world contexts. This study does not claim to present a final or definitive solution; rather, it represents a step toward assessing the feasibility of indirect and behavior-based approaches in psychological evaluation. Moreover, the purpose of these tools is to support individuals in identifying career options that align with their current characteristics, not to enforce or prescribe specific career decisions.

\section{Conclusion }
This study investigated the feasibility of using serious-game behavioral data to predict suitability for software development roles as an alternative to traditional questionnaire-based personality assessment. By combining insights from the Myers-Briggs framework with empirically validated developer-specific behavioral traits, a multi-stage game was designed to elicit meaningful cognitive and behavioral signals under controlled yet engaging conditions. The experimental results provide strong evidence that gameplay-derived behavioral features can serve as reliable predictors of software-development suitability, achieving high accuracy and precision even in the absence of explicit personality data. From a practical perspective, the proposed framework can support early-stage career guidance, educational counseling, and exploratory screening by providing engaging, low-pressure assessments that respect user autonomy and do not prescribe career outcomes. The study does not claim to deliver a final or clinically validated selection tool; rather, it establishes a methodological foundation and empirical evidence for behavior-based assessment using games.

\textbf{Disclosure statement}

The authors declare that there is no conflict of interest regarding the publication of this paper.

\textbf{Funding}

This research did not receive any specific grant from funding agencies in the public, commercial, or not-for-profit sectors.

\bibliography{report} 
\bibliographystyle{spiebib} 

\clearpage

\appendix    

\section{Game Design, Mechanics, and Data Collection Details}
\label{sec:misc}

This appendix provides a detailed description of the serious games developed for behavioral and personality assessment, including their mechanics, stages, and the data acquisition process.

\subsection{Group-Swapping Puzzle Game}
The first game is a puzzle in which players must swap the positions of two groups of pieces using a limited number of moves within a constrained grid. Movements are only allowed to valid, unoccupied cells and are executed via keyboard input after selecting a piece. Invalid moves are ignored and not counted.

Each level has an optimal solution in terms of time and number of moves. If the time limit expires, players may surrender or continue playing. The game consists of three stages: an introductory tutorial stage, a medium-difficulty stage, and a high-difficulty stage. Figure 8 illustrates the interface of the group-swapping puzzle game used in the experiment.

\begin{figure}[!htbp]
\centering
\includegraphics[width=1\linewidth]{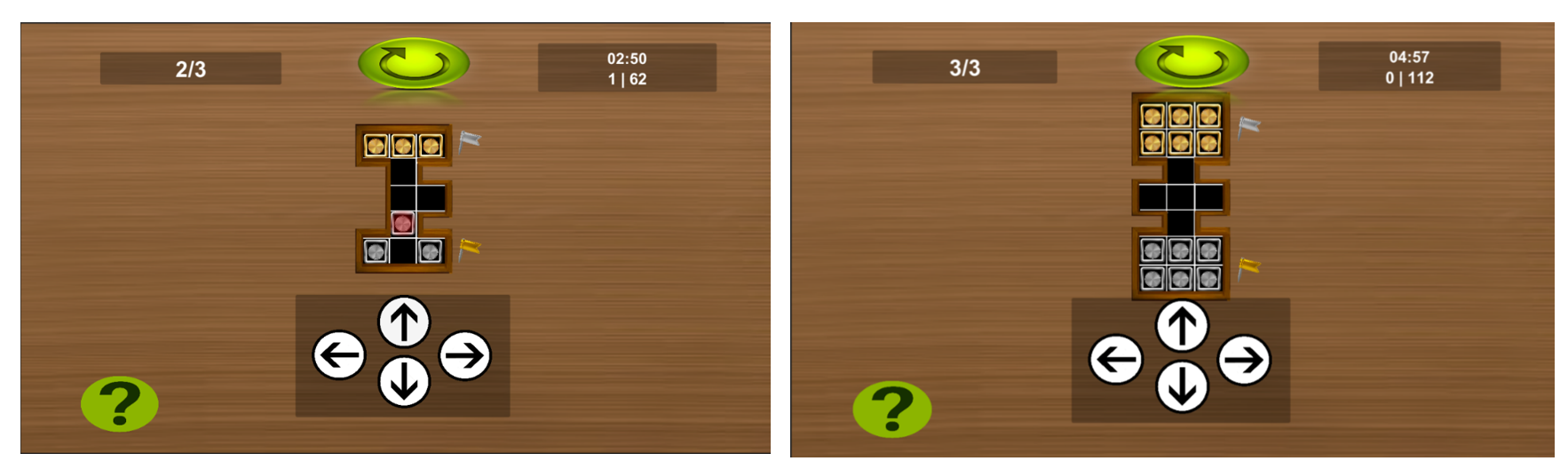}
\caption{
Group-swapping puzzle game interface
}
\label{fig:example}
\end{figure}

To reduce frustration, players are allowed to use a limited number of “skip tokens” to bypass puzzle stages without penalty. Initially, three skips are provided, and additional skips can be earned by completing optional logical or mathematical challenges. A sample of skip-token and optional challenge mechanism is presented in Figure 9.

\begin{figure}[!htbp]
\centering
\includegraphics[width=1\linewidth]{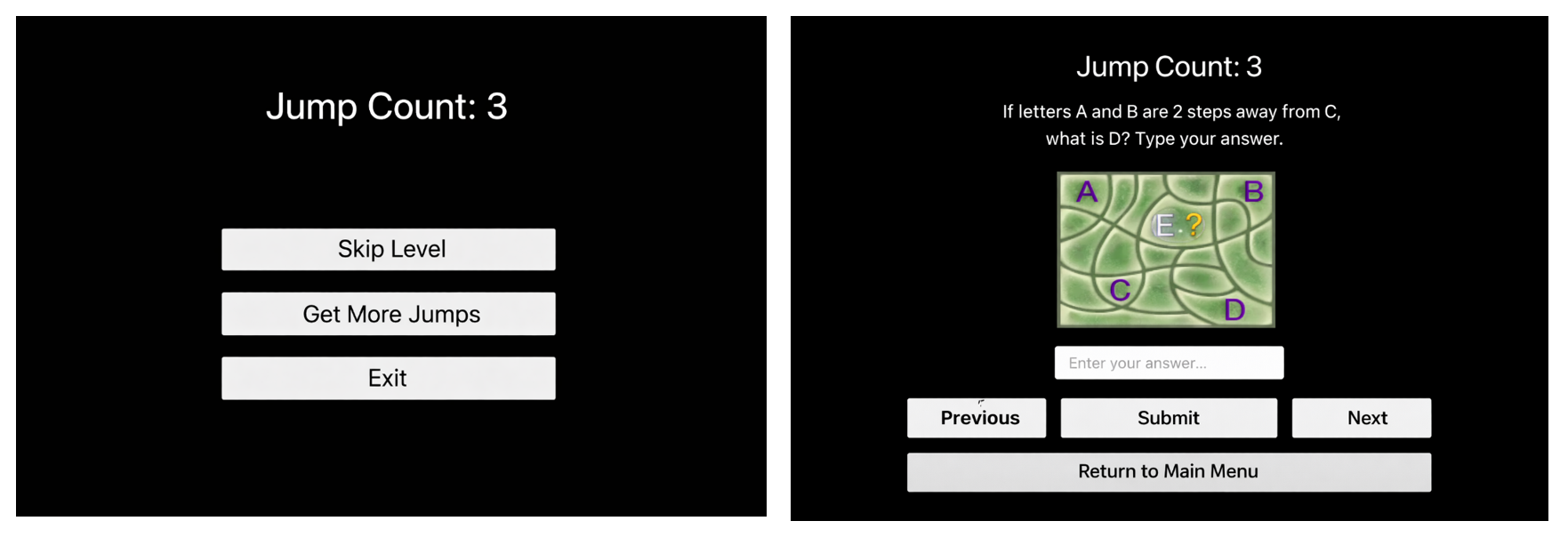}
\caption{
Skip-token and optional challenge mechanism
}
\label{fig:example}
\end{figure}

\subsection{Memory Matching Game}
The memory game presents pairs of numbers to the player for five seconds. Afterward, the cards are flipped, and the player must identify matching pairs from memory. If the player fails, repeated attempts are allowed, encouraging efficient memorization strategies.

This game includes three difficulty levels, increasing the number of cards progressively. No skip option is provided in this game to ensure full engagement with memory-based behavior. Figure 10 illustrates the interface of the memory-matching game used in the experiment

\begin{figure}[!htbp]
\centering
\includegraphics[width=1\linewidth]{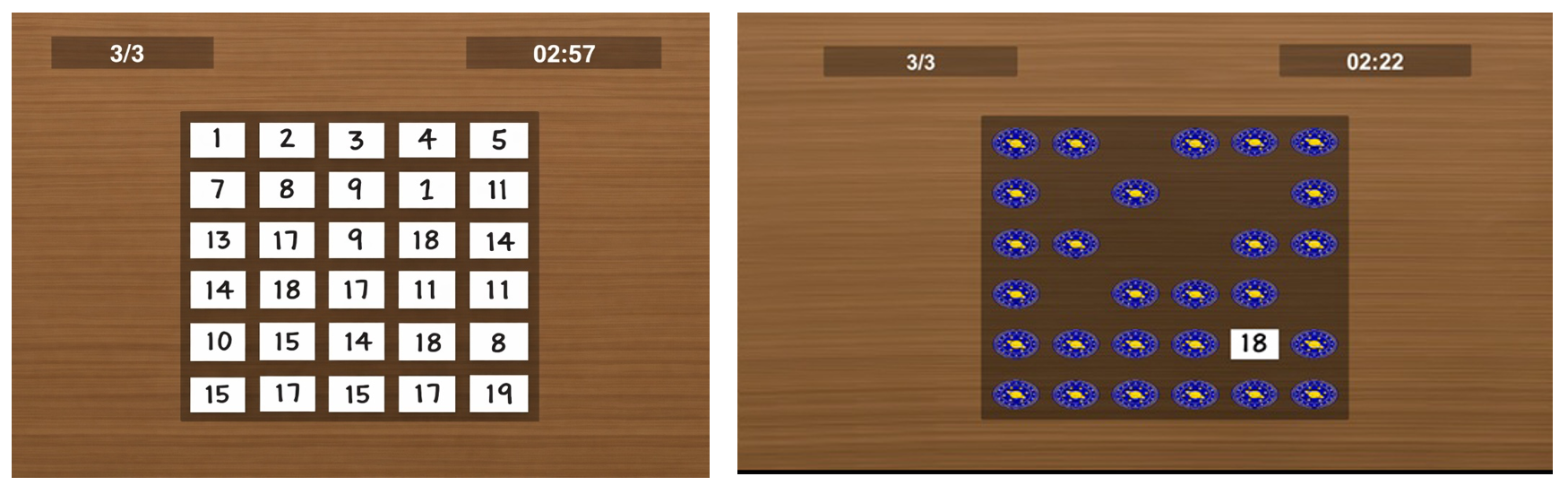}
\caption{
Memory-matching game interface
}
\label{fig:example}
\end{figure}

\subsection{Galaxy Shooter Game}
The galaxy shooter game is an action-based shooter where players must survive for 120 seconds while encountering different enemy types, including offensive, defensive, and score-based entities. Optional side challenges are available in the menu. Samples of tutorial interfaces of the shooter game are shown in Figure 11.

\begin{figure}[!htbp]
\centering
\includegraphics[width=1\linewidth]{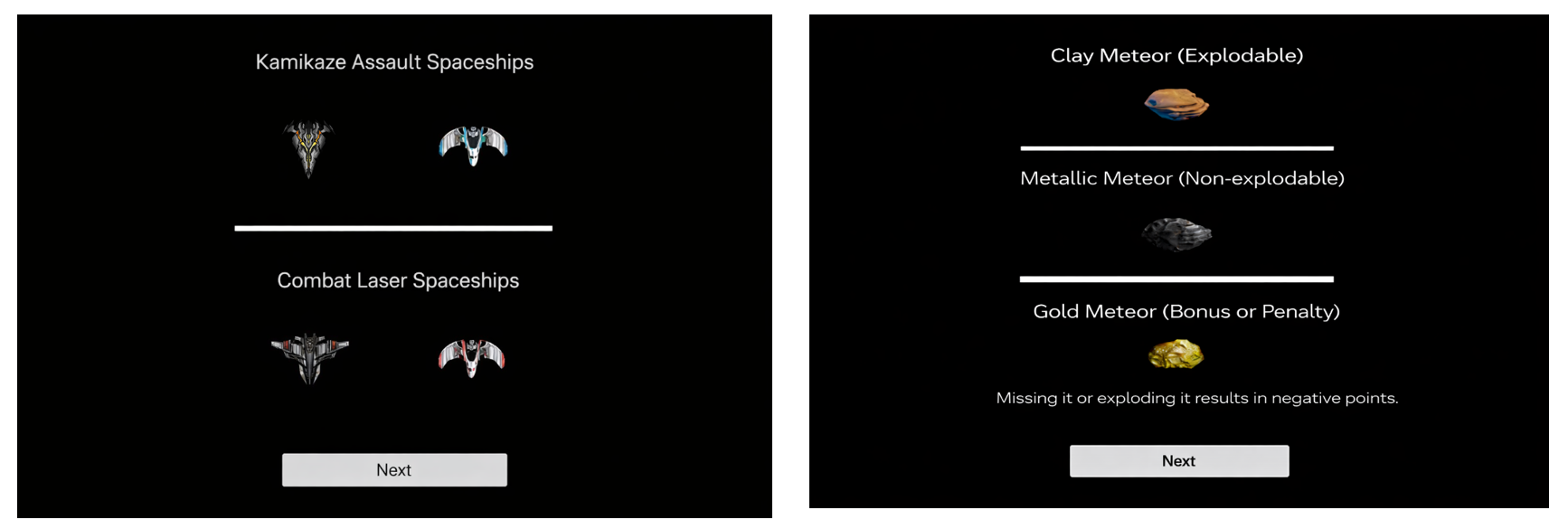
}
\caption{
Tutorial interfaces in the shooter game
}
\label{fig:example}
\end{figure}

After three failed attempts, players may surrender and proceed. Multiple tutorial screens explain controls and enemy behavior. This game primarily captures time management, reaction speed, and stress-related decision-making behavior. The galaxy shooter game interface is shown in Figure 12.

\begin{figure}[!htbp]
\centering
\includegraphics[width=0.25\linewidth]{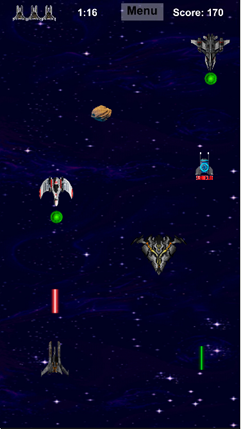
}
\caption{
Galaxy shooter game interface
}
\label{fig:example}
\end{figure}

\subsection{Obstacle-Rearrangement Path Game}
In this puzzle, players must move obstacles to guide a target piece to a designated endpoint. Obstacles have constrained movement patterns and include single-cell blocks, two-cell vertical and horizontal blocks, and square-shaped blocks (as shown in Figure 13). The game includes three stages with increasing complexity and limited movement freedom.

\begin{figure}[!htbp]
\centering
\includegraphics[width=0.5\linewidth]{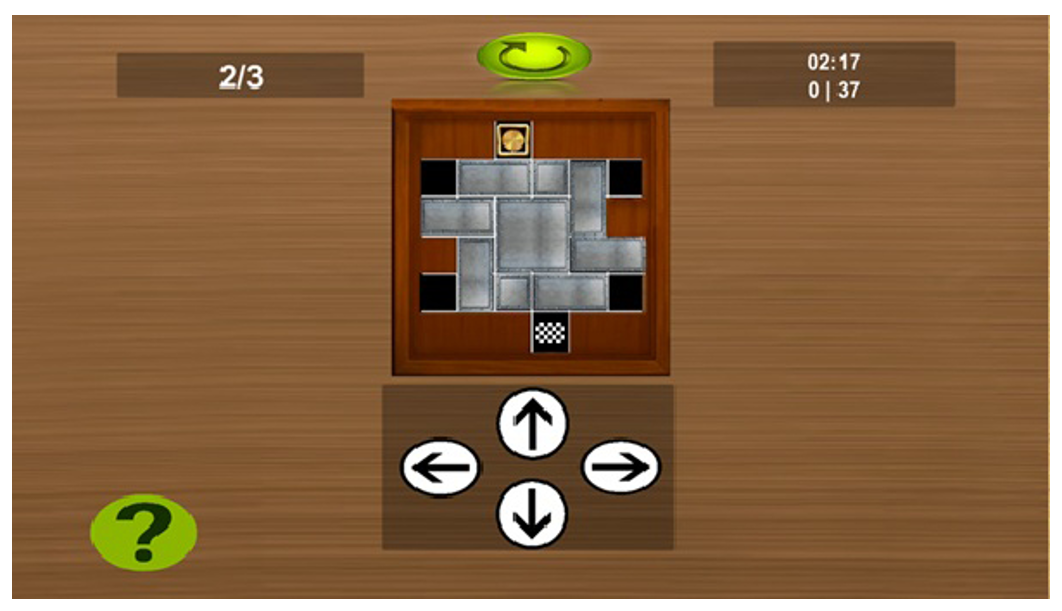
}
\caption{
Obstacle-rearrangement path game interface
}
\label{fig:example}
\end{figure}

\subsection{Graph Traversal Game}

This game requires players to traverse all vertices of a graph without revisiting any node. Movement continues automatically in the chosen direction until a previously visited node or obstacle is encountered. Due to player feedback, the original complex rules were simplified, and the final version consists of four progressively challenging stages. Figure 14 illustrates the interface of the graph traversal game.

\begin{figure}[!htbp]
\centering
\includegraphics[width=0.5\linewidth]{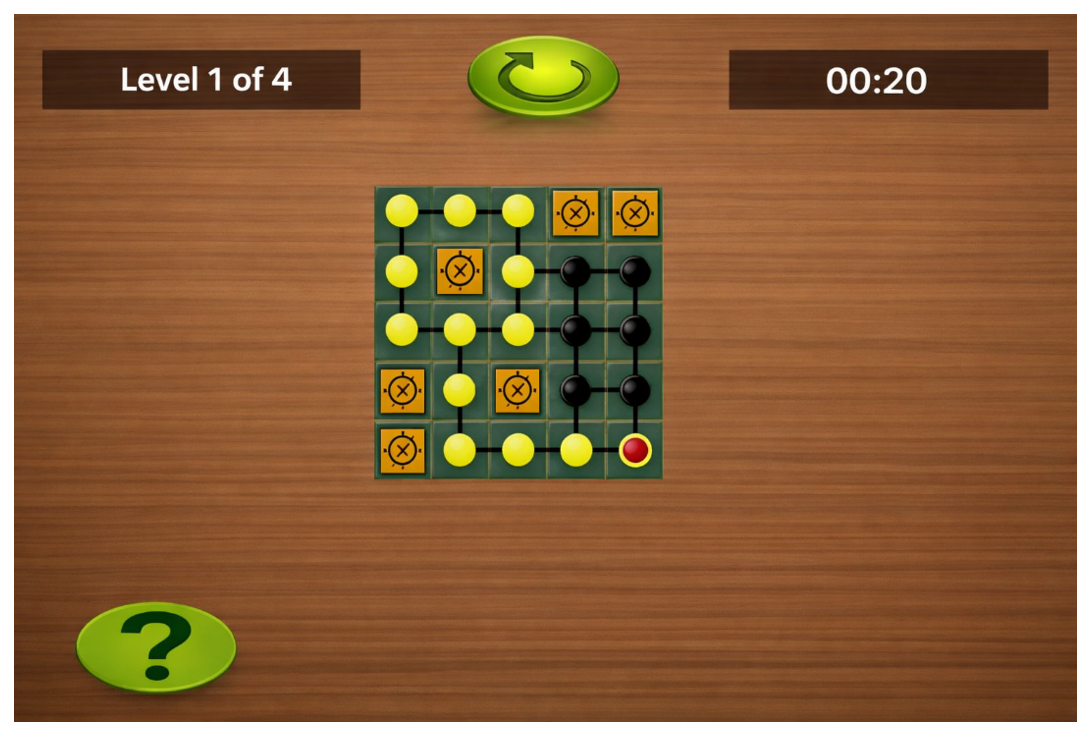
}
\caption{
Graph traversal game interface
}
\label{fig:example}
\end{figure}

\subsection{Gameplay Logging and Data Submission}
Upon completing the games, players were asked to submit their gameplay logs to the server (Figure 15). A unique five-digit tracking code was generated for each participant, corresponding to the final digits of a hash derived from their gameplay logs. This code enabled anonymized tracking and reliable data association during analysis.

\begin{figure}[!htbp]
\centering
\includegraphics[width=0.5\linewidth]{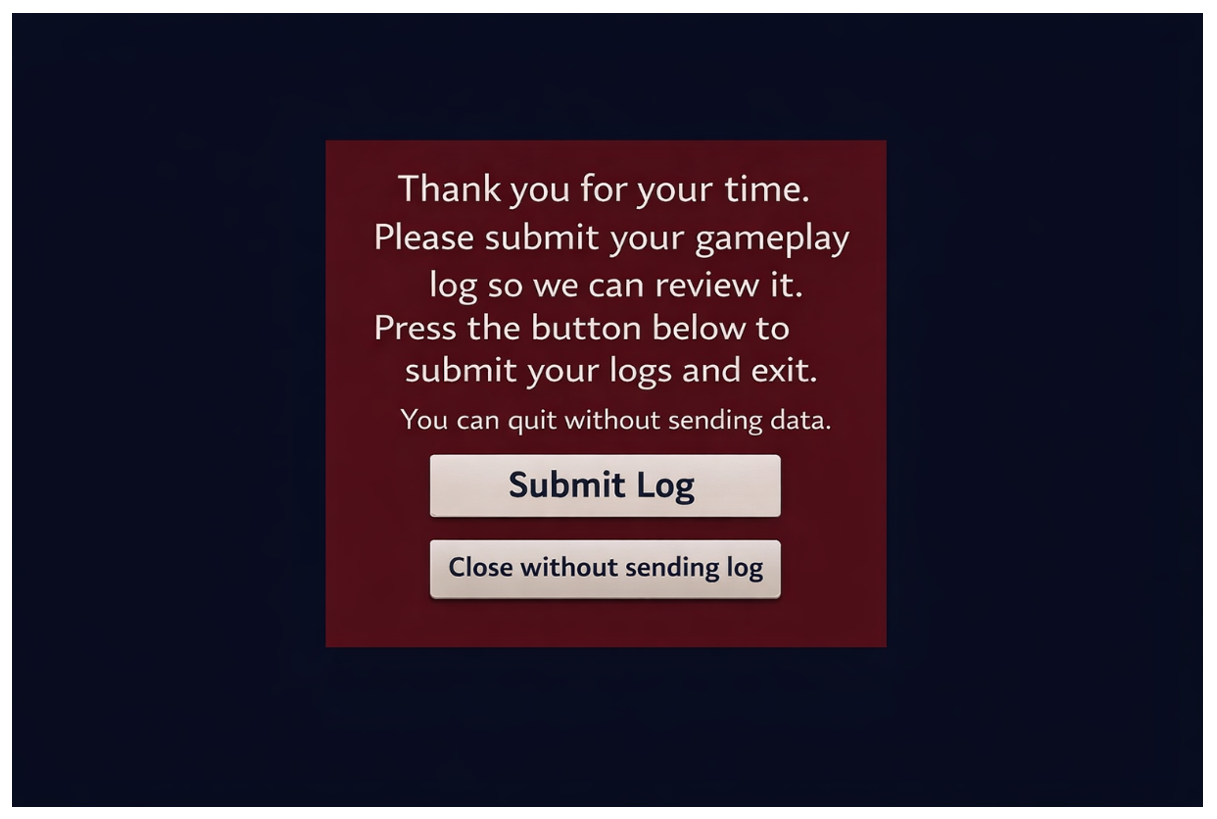
}
\caption{
Gameplay data submission and tracking interface
}
\label{fig:example}
\end{figure}

\subsection{Excluded Prototype Games}

During beta testing, additional games were evaluated but excluded from the final version. These included a narrative card-based role-playing game intended to assess personality through storytelling choices, which was removed due to low engagement. A tower defense game focusing on strategic planning and resource management was also excluded because of excessive complexity and insufficient learning time for participants. These two games showed in Figure 16.

\begin{figure}[!htbp]
\centering
\includegraphics[width=1\linewidth]{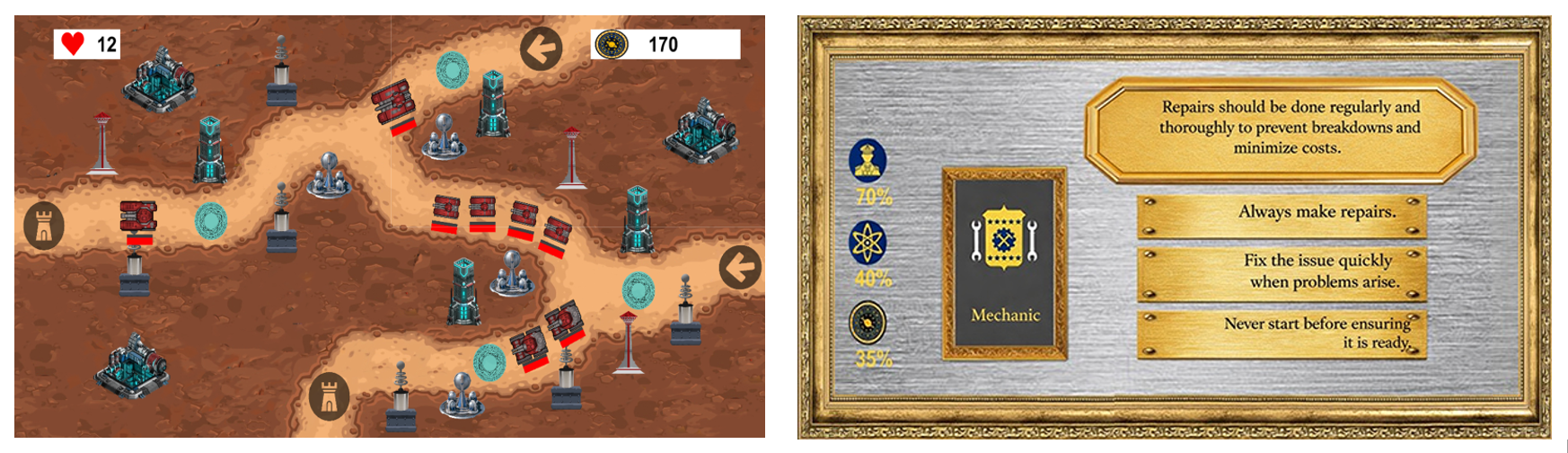
}
\caption{
Excluded prototype game designs
}
\label{fig:example}
\end{figure}

\clearpage

\section{List of Variables}
Unique Gameplay Tracking Code (String) - Participant Age (Number) - Participant Gender (Enum) - Gender Flag: Male (Bool) - Gender Flag: Female (Bool) - Computer Engineering Background (Bool) - MBTI Personality Type (Enum) - MBTI Extraversion-Introversion (Enum) - MBTI Sensing-Intuition (Enum) - MBTI Thinking-Feeling (Enum) - MBTI Judging-Perceiving (Enum) - MBTI SJ Functional Group (Bool) - MBTI SP Functional Group (Bool) - MBTI NF Functional Group (Bool) - MBTI NT Functional Group (Bool) - Confirmed Programming Interest (Bool) - Predicted Programming Suitability (Bool) - Help-Seeking Behavior (Bool) - Problem-Solving via Online Search Skill (Bool) - Time Management Ability (Bool) - Competitive Motivation (Bool) - Behavioral Flexibility and Adaptability (Bool) - Primary Gaming Platform (Enum) - Average Weekly Gameplay Duration (Number) - Selected Game Difficulty Level (Enum) - Total Gameplay Log Count (Number) - Total Gameplay Duration in Minutes (Number) - Total Gameplay Pause Count (Number) - Total Game Restart Count (Number) - Group-Swapping Puzzle Game: Log Count (Number) - Group-Swapping Puzzle Game: Gameplay Duration in Minutes (Number) - Group-Swapping Puzzle Game: Skip Token Usage Count (Number) - Group-Swapping Puzzle Game: Surrender Action Count (Number) - Obstacle-Rearrangement Path Game: Log Count (Number) - Obstacle-Rearrangement Path Game: Gameplay Duration in Minutes (Number) - Memory Matching Game: Log Count (Number) - Memory Matching Game: Gameplay Duration in Minutes (Number) - Memory Matching Game: Total Guess Count (Number) - Memory Matching Game: Correct Guess Count (Number) - Memory Matching Game: Incorrect Guess Count (Number) - Tutorial Engagement Count (Number) - Tutorial Skipping Count (Number) - Quiz Navigation Interaction Count (Number) - Galaxy Shooter Game: Gameplay Log Count (Number) - Galaxy Shooter Game: Maximum Score Achieved (Number) - Galaxy Shooter Game: Total Shots Fired (Number) - Galaxy Shooter Game: Total Lives Lost (Number) - Galaxy Shooter Game: Restart Count (Number) - Galaxy Shooter Game: Power-Ups Generated (Number) - Galaxy Shooter Game: Power-Ups Collected (Number) - Galaxy Shooter Game: Gold Generated (Number) - Galaxy Shooter Game: Gold Collected (Number) - Galaxy Shooter Game: Gold Lost (Number) - Galaxy Shooter Game: Gold Exploded (Number) - Galaxy Shooter Game: Enemies Generated (Number) - Galaxy Shooter Game: Enemies Destroyed by Shooting (Number) - Galaxy Shooter Game: Enemy Collisions with Player (Number) - Galaxy Shooter Game: Asteroids Generated (Number) - Galaxy Shooter Game: Asteroids Destroyed by Shooting (Number) - Galaxy Shooter Game: Asteroid Collisions with Player (Number) - Galaxy Shooter Game: Leftward Movement Count (Number) - Galaxy Shooter Game: Rightward Movement Count (Number) - Galaxy Shooter Game: Total Horizontal Movement Count (Number) - Galaxy Shooter Game: Left Boundary Exit Count (Number) - Galaxy Shooter Game: Right Boundary Exit Count (Number) - Galaxy Shooter Game: Total Boundary Exit Count (Number) - Galaxy Shooter Game: Life Survival Challenge Completed (Bool) - Galaxy Shooter Game: Enemy Elimination Challenge Completed (Bool) - Galaxy Shooter Game: Asteroid Destruction Challenge Completed (Bool) - Galaxy Shooter Game: Gold Collection Challenge Completed (Bool) - Galaxy Shooter Game: No-Weapon Usage Challenge Completed (Bool) - Galaxy Shooter Game: Score Achievement Challenge Completed (Bool) - Graph Traversal Game: Log Count (Number) - Graph Traversal Game: Gameplay Duration in Minutes (Number) - Galaxy Shooter Game: Gameplay Logs per Minute (Number) - Galaxy Shooter Game: Wins per Minute (Number) - Galaxy Shooter Game: Shots per Minute (Number) - Galaxy Shooter Game: Shots per Gameplay Log (Number) - Galaxy Shooter Game: Lives Lost per Minute (Number) - Galaxy Shooter Game: Lives Lost per Gameplay Log (Number) - Galaxy Shooter Game: Power-Up Collection Efficiency Ratio (Number) - Galaxy Shooter Game: Gold Collection Efficiency Ratio (Number) - Galaxy Shooter Game: Enemy Collision Rate (Number) - Galaxy Shooter Game: Asteroid Collision Rate (Number) - Galaxy Shooter Game: Directional Movement Bias Ratio (Number) - Galaxy Shooter Game: Boundary Exit Bias Ratio (Number)

\end{document}